\documentclass[11pt]{mystyle}

\usepackage[numbers,sort&compress]{natbib}
\usepackage{fontawesome5}
\AddToHook{package/axessibility/after}{\pdfminorversion=7\relax}
\usepackage{multirow}
\usepackage{boldline}
\usepackage{pifont}
\newcommand{\benchpartial}{\mbox{\usefont{U}{wasy}{m}{n}\rlap{\char71}\char35}}
\usepackage{amssymb}
\usepackage{bm}

\usepackage{threeparttable}
\usepackage{tabularx}
\usepackage[accsupp]{axessibility}  
\definecolor{Bpurple}{HTML}{BF40BF}

\usepackage{colortbl}
\usepackage{float}

\usepackage{wrapfig,placeins}

\microtypesetup{expansion=true}
\newcommand{\proselines}{\setlength{\parfillskip}{0pt plus .30\linewidth}}
\appto{\absfont}{\proselines}
\DeclareCaptionJustification{proselines}{\proselines}
\makeatletter
\apptocmd{\ps@firststyle}{}{}{}
\makeatother
\definecolor{WorldLeagueTitleBackground}{HTML}{FFF0D6}
\tcbset{titlebox/.append style={colback=WorldLeagueTitleBackground,left=4mm,right=4mm}}
\newsavebox{\projectrowsbox}
\appto{\abscontent}{%
  \par\medskip
  {\normalfont\footnotesize
  \sbox{\projectrowsbox}{%
    \begin{tabular}{@{}>{\centering\arraybackslash}m{1.55em}@{\hspace{0.2em}}>{\sffamily\bfseries}l@{\quad}l@{}}
      \faGlobe & Homepage: & \url{https://xyangzhou.github.io/WorldLeague} \\
      \raisebox{-0.15\height}{\includegraphics[height=1.18em]{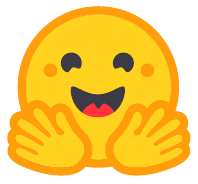}}
        & Hugging Face: & \url{https://huggingface.co/datasets/xyangzhou/WorldLeague} \\
      \faGithub & GitHub: & \url{https://github.com/xyangzhou/WorldLeague} \\
      \faEnvelope & Email: & \{\href{mailto:xinyangzhou@u.nus.edu}{xinyangzhou}, \href{mailto:liqi@u.nus.edu}{liqi}\}@u.nus.edu, \href{mailto:xinchao@nus.edu.sg}{xinchao}@nus.edu.sg
    \end{tabular}%
  }%
  \noindent\usebox{\projectrowsbox}\hfill
  \raisebox{\dimexpr(\ht\projectrowsbox-\dp\projectrowsbox-\height)/2\relax}{%
    \includegraphics[height=36pt]{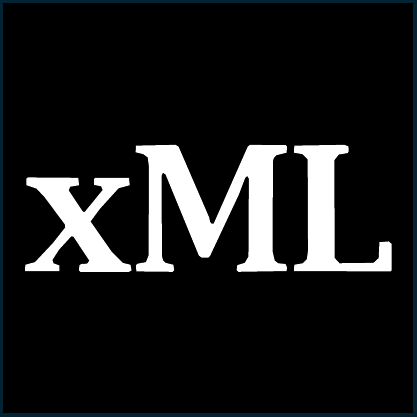}%
  }\kern 4pt\par}%
}
\hypersetup{
  colorlinks=true,
  allcolors=black,
  urlcolor=Bpurple,
  filecolor=Bpurple,
  pdftitle={WorldLeague Benchmarking Physical Agency}
}

\runningtitle{\includegraphics[height=18pt]{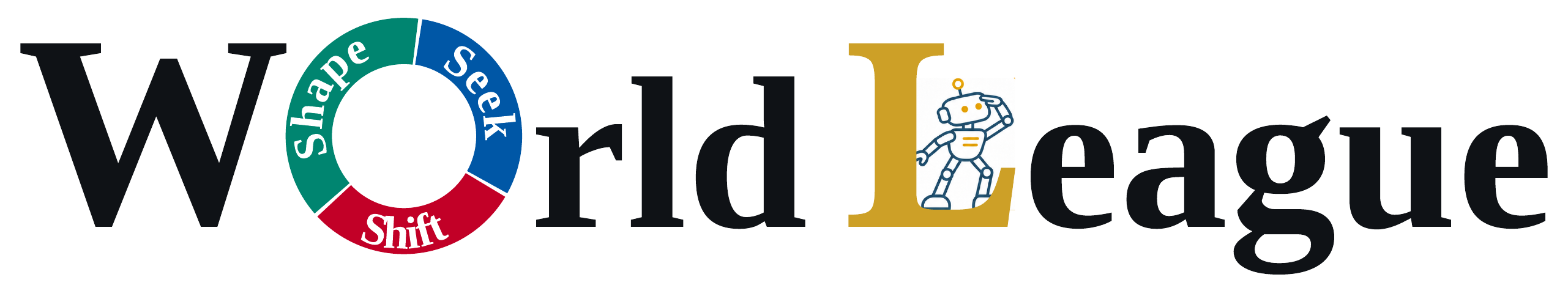}}
\title{%
  \includegraphics[width=0.45\linewidth]{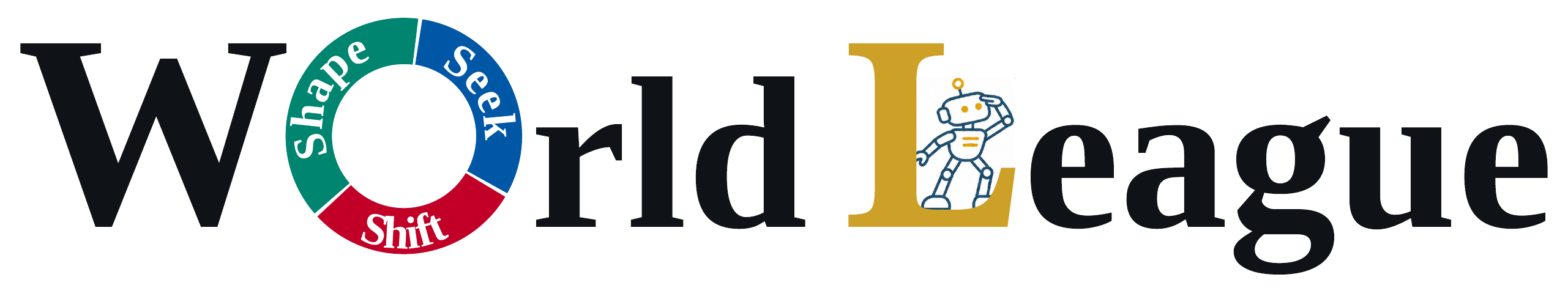}\\[-2pt]
  {\fontsize{12}{13}\selectfont Benchmarking Physical Agency}}

\renewcommand{\Affilfont}{\centering\normalfont\fontsize{8}{10}\selectfont}
\author{Xinyang Zhou, Qi Li, Xinchao Wang\protect\footnotemark[2]\\
xML Lab, National University of Singapore}
\date{}

\begin{document}
\begin{abstract}
AI systems can now operate Blender, follow instructions in complex video games, and manipulate objects with robot arms.
What range of physical goals can today's AI systems achieve autonomously?
Answering this question requires testing systems on tasks that demand different combinations of perception, memory, planning, and action.
We introduce \textbf{WorldLeague}, a benchmark that compares AI systems on a common set of \textbf{60 simulated robot tasks} by measuring progress toward each task's physical goal from the robot's trajectory.
Robots explore scenes to gather evidence in \textbf{WorldSeek}, rearrange objects in \textbf{WorldShape}, and adapt to changes in \textbf{WorldShift}.
Tying these goals to a single control interface limits which AI systems can be compared.
In WorldLeague, vision-language models (VLMs) choose targets and motions, while vision-language-action models (VLAs) and world-action models (WAMs) use their native controls.
For each of six capability dimensions, we measure how much progress a system makes toward its goals (Interactive Score) and how often it achieves them (physical success rate).
Together, these measures form the system's \emph{Physical Agency Profile}.
Among 13 representative AI systems, Claude Fable 5.2 achieves 13 of the 60 goals and earns the highest mean Interactive Score (28.58/100).
Fable leads in four dimensions, while GPT-6.1-Sol and GPT-6-Astra tie for first on rule-following tasks.
Even when VLMs answer every final question correctly, they receive an Interactive Score of zero in 71.8\% of those episodes.
WorldLeague gives Embodied AI, Physical AI, and Agentic AI a shared basis for comparing what AI systems achieve autonomously, from partial progress toward a physical goal to full goal completion.
\end{abstract}
\maketitle

\footnotetext[2]{Corresponding author.}

\par\smallskip
\noindent\begin{minipage}{\linewidth}
  \centering
  \captionsetup{type=figure,skip=4pt,font=footnotesize}
  \includegraphics[width=0.88\linewidth]{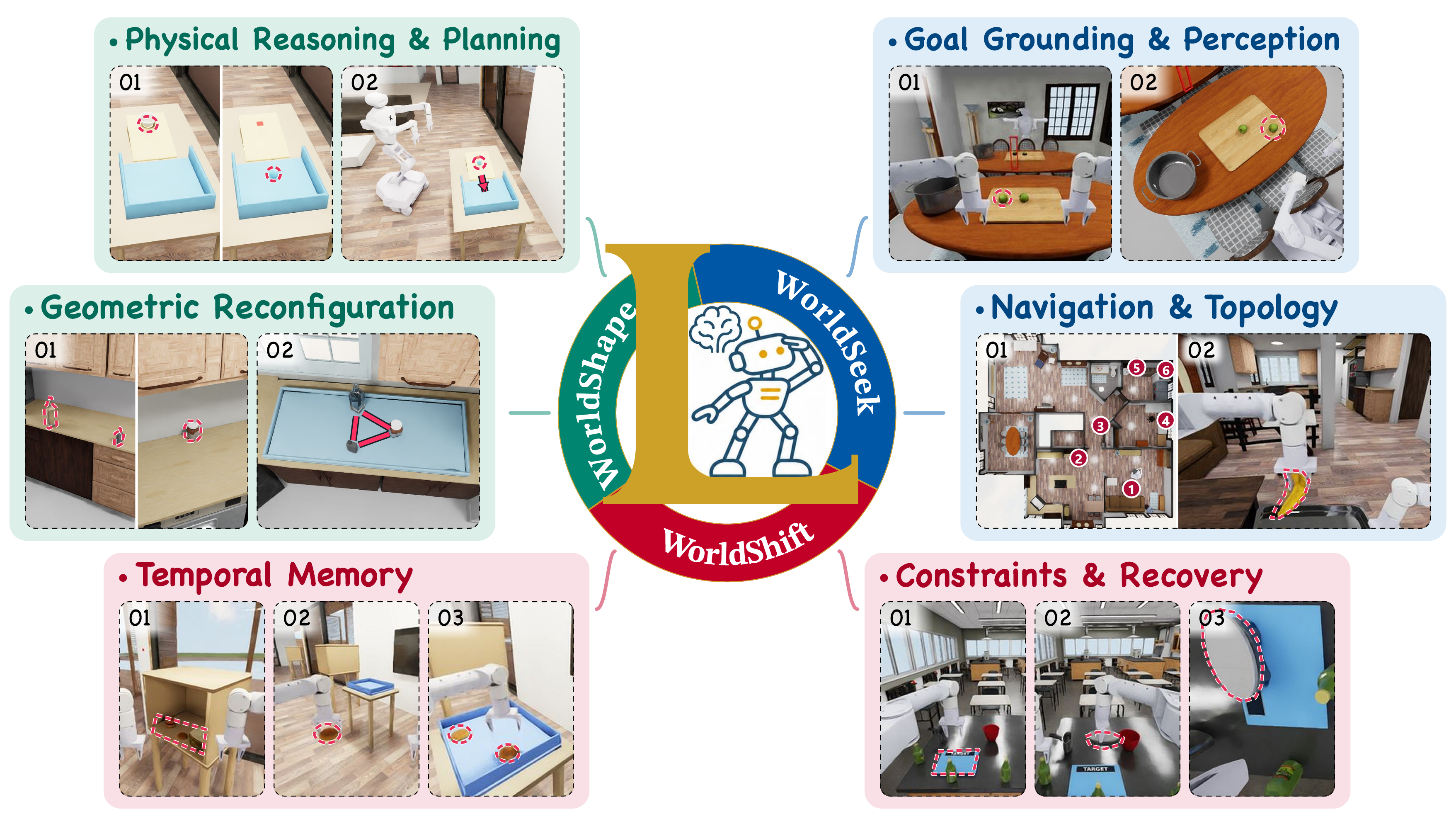}
  \caption{Example tasks across six dimensions in WorldSeek (blue, right), WorldShape (green, left), and WorldShift (red, below).}
  \label{fig:worldleague_overview}
\end{minipage}\par

\proselines
\setlength{\parskip}{0.62\baselineskip}
\section{Introduction}
\label{sec:introduction}

Foundation models have recently become a promising direction for building general-purpose robots~\cite{team2024octo}.
In digital environments, agents built on frontier models can follow instructions in open-world video games~\cite{bolton2025sima} and turn visual references into editable 3D scenes in Blender~\cite{fang2026blenderworld}.
The range of tasks that physical robots can perform has also grown~\cite{han2026video2sim2real,liu2026egoengine}.
Robots have performed long-horizon household tasks, such as cleaning a kitchen or bedroom, in homes they had never seen during training~\cite{intelligence2025pi}.

To clean an unfamiliar kitchen, a robot must use what it sees to decide what to do next. Embodied AI systems connect perception to action in different ways~\cite{huang2026best}. A vision-language model (VLM) can identify an object from a spatial description~\cite{chen2026behavior} and estimate its distance from other objects~\cite{chen2024spatialvlm}. With a robot interface, it can use such judgments to choose where to move or grasp~\cite{chen2026show}. Vision-language-action (VLA) models produce robot actions directly from instructions and visual observations~\cite{kim2024openvla}. For example, Gemini Robotics 1.5 interleaves reasoning and action during multi-step manipulation~\cite{team2025gemini}. World-action models (WAMs) jointly predict future observations and robot actions~\cite{li2026causal,shen2026world,wang2026openwam,li2026badwam}. This approach has enabled closed-loop manipulation on new tasks and in new environments~\cite{ye2026world}.

Across these approaches, the ambition is the same. A person specifies a physical goal and leaves the system to decide how to achieve it~\cite{liu2023agentbench,chen2026show,li2026vision,li2026sponge,qian2025wristworld,robotics2026envisioning}. We call the system making those decisions the \emph{agent} and its capacity to choose and execute actions through a body its \emph{physical agency}. The decisions an agent must make depend on the goal. To find a hidden object, it must decide where to look and what to move out of the way~\cite{das2018embodied,danielczuk2019mechanical,huang2026era}. To put that object through a narrow opening, it must also choose an orientation that fits.
An agent may therefore find the right object and still fail to put it away.
This leads us to the central question of our study.
\emph{What range of physical goals can today's AI systems achieve autonomously?}

Organizing robot benchmarks around household activities~\cite{li2023behavior,nasiriany2026robocasa365} obscures what success on each activity establishes about an agent. Consider a robot asked to deliver a cup. If the user names the cup and it is in plain sight, the robot can select it from its current view. If the user asks for the cup that has moved since the robot last visited the room, the robot must use an earlier observation to decide what to pick up. Both tasks count as cup delivery. Yet success on the first leaves the decisive requirement of the second entirely untested. To establish the range of an agent's physical abilities, we must vary what the agent has to work out for itself and verify that its actions bring about the requested outcome.

We introduce \textbf{WorldLeague}, a benchmark of \textbf{60 simulated robot tasks} organized by the physical demands that determine success.
We evaluate VLMs, VLAs, and WAMs against shared goals and constraints.
For each task, the robot's trajectory shows which goal requirements it satisfies and whether it completes the task.
The tasks span three tracks, each defined by what the robot must do to achieve its goal.
\textbf{WorldSeek} tests navigation and goal grounding as systems explore and gather evidence. \textbf{WorldShape} tests geometric reconfiguration and physical reasoning as systems change the scene. \textbf{WorldShift} tests temporal memory and constraint management as systems track a changing world and recover from disturbances. The three tracks span six capability dimensions (Sec.~\ref{sec:world_challenges}). We group each task by its primary challenge, and a single task can draw on capabilities from several dimensions.

To compare systems across these tasks, we instantiate each goal in a scene with specified objects, initial conditions, and constraints. Physical predicates check conditions such as stable placement or recovery after a disturbance. We inspect all 60 task--scene pairs, verify that reference runs meet the goals and constraints, and check the scoring conditions against the task specifications.
Each system pursues its goal through physical interaction. For VLMs, our \emph{decision-preserving action abstraction} draws on how sensorimotor mechanisms coordinate fine-grained human movement~\cite{todorov2002optimal, wolpert2011principles, graybiel2008habits}. The interface executes motions, and the agent chooses the targets, action parameters, sequences, and responses to feedback that shape the outcome. We adapt a motion interface~\cite{chen2026show} to implement this abstraction on R1~\cite{li2023behavior}. VLA and WAM policies use their native interfaces under shared task goals, constraints, and physical scoring rules.

We score the resulting trajectory for progress toward individual requirements and for physical success, which requires satisfying the full goal and all task constraints. A \emph{Physical Agency Profile} reports Interactive Score for progress and physical success rate for complete goal achievement in each capability dimension. For VLMs, terminal question answering (QA) provides an auxiliary measure of scene understanding, outcome recognition, and memory within the same episodes.

Among the 13 evaluated embodied AI systems, Claude Fable 5.2 earns the highest overall Interactive Score, 28.58 out of 100, and achieves physical success on 13 of its 60 tasks. Fable leads in navigation, goal grounding, geometric reconfiguration, and temporal memory, while GPT-6.1-Sol and GPT-6-Astra lead in constraint management. The trajectory analysis also identifies goal requirements that remain unmet after systems make physical progress. Of 347 VLM episodes with full terminal QA credit, 249 (71.8\%) receive an Interactive Score of zero. These results show that goal-relevant understanding and successful intermediate actions can coexist with unmet physical goals.

Turning demonstrated capabilities into desired physical outcomes is central to Embodied AI, Physical AI, and Agentic AI, which share the ambition of building systems that can pursue human goals through their own actions in the world. WorldLeague turns this ambition into a common experiment. A person specifies the desired outcome, an agent determines how to pursue it through a body, and the evaluation verifies what it achieves in the physical environment. Evaluating systems against common goals and physical success criteria allows advances in perception, reasoning, and control to be assessed in terms of what they enable these systems to accomplish~\cite{wang2026practical}. Our broader aim is to make autonomous goal achievement a shared measure of progress for AI systems that act in the world.

We make three contributions to the evaluation of physical agency in embodied AI systems.
\begin{itemize}
    \item We test physical agency by goal achievement on tasks with different capability combinations.
    \item We build and validate WorldLeague, a suite of 60 tasks that measures progress and physical success across six capability dimensions using shared goals and scoring rules.
    \item We compare the capability profiles of 13 systems and identify goal requirements that remain unmet in episodes with correct answers or partial physical progress.
\end{itemize}
\section{Related Work}
\label{sec:related_work}

\subsection{The Demands of Physical Goals}

BEHAVIOR-1K~\cite{li2023behavior} takes everyday household activities as the unit of evaluation. Within a single activity, what a robot must decide can change as it acts~\cite{liu2023reflect}. In RoboQuest~\cite{renhang2026roboquest}, a robot uses trial impressions to decide how to stamp the requested pattern. MeMoREval~\cite{wang2026should} makes a later request depend on what the system encountered earlier, while RobotAgentBench~\cite{mao2026robotagentbench} introduces disturbances that require an agent to revise an ongoing manipulation. WorldLeague organizes goals by the primary challenge they pose. Finding a specified object and fitting it through a narrow opening both require identifying the object, but the second goal also requires reasoning about orientation and clearance. Grouping tasks this way lets us trace how success varies across exploration, physical transformation, and adaptation.

\subsection{Action Abstraction for Embodied AI Systems}

An action interface determines which decisions the agent makes~\cite{yang2025embodiedbench}. SayCan~\cite{ahn2022can} uses a skill's feasibility to guide selection, and Code as Policies~\cite{liang2023code} lets a model compose robot actions in a program. Inner Monologue~\cite{huang2022inner} feeds execution outcomes back into planning. Show-Harness~\cite{chen2026show} translates agent-selected commands through an embodiment-specific interpreter, and HumanCLAW~\cite{siyao2026humanclaw} uses the separation between action selection and motor execution to study how VLMs direct a body. WorldHarness adapts Show-Harness to preserve the VLM's decisions throughout a task. To place an object, the VLM chooses where to grasp, how far to move, and when to release. The controller executes the motion and returns feedback that the VLM can use to correct its next move. VLA and world-action policies pursue the same goals through their native interfaces.

\subsection{Measuring Physical Goal Achievement}

ESI-Bench~\cite{closesesi} lets systems move and manipulate objects to gather evidence for a spatial answer. ALFRED~\cite{shridhar2020alfred} scores both task success and the fraction of goal conditions met. IMBench~\cite{maurya2026imbench} examines understanding, planning, and execution in separate stages. WorldLeague builds on selected ESI-Bench scene records to pair what a VLM can report with what it physically achieves in the same episode. Once the physical outcome is fixed, terminal QA probes the VLM's understanding of the scene and its actions. Across all three model families, Physical Agency Profiles show how partial progress and full goal achievement vary across the six dimensions.

\subsection{Comparison with Existing Benchmarks}
\label{sec:comparison_synthesis}

Table~\ref{tab:benchmark_comparison} compares 17 benchmarks by the capabilities their tasks score and the actions their agents control (Appendix~\ref{sec:comparison_criteria}). RoboQuest and IMBench couple active perception to manipulation by asking agents to uncover information they need to change a physical configuration. MeMoREval tests whether an earlier observation guides a later choice, and RobotAgentBench tests recovery under constraints. WorldLeague evaluates all six dimensions through physical goal achievement, so the same system's ability to find and manipulate a target can be compared with its ability to track that target as the scene changes.

\begin{table}[!htbp]
\centering
\caption{Benchmark coverage of WorldLeague's six dimensions (Table~\ref{tab:task_taxonomy}). Dimension marks denote joint tests (\ding{51}), component tests (\benchpartial), or no test (\ding{55}). In the Execution column, the same marks denote feedback-guided robot motion, delegated skills/symbolic actions/active QA, and passive QA/prediction, respectively.}
\label{tab:benchmark_comparison}
\begingroup
\footnotesize
\setlength{\tabcolsep}{3pt}
\renewcommand{\arraystretch}{1.18}
\renewcommand{\tabularxcolumn}[1]{m{#1}}
\newcommand{\bcY}{{\normalsize\ding{51}}}
\newcommand{\bcP}{{\normalsize\benchpartial}}
\newcommand{\bcN}{{\normalsize\ding{55}}}
\newcommand{\bcHeader}[5]{%
  \multicolumn{#1}{>{\columncolor[HTML]{#2}[#3][#4]}c}{#5}}
\newcommand{\bcExecutionHeader}[1]{%
  \multicolumn{1}{>{\columncolor[HTML]{EBEDF3}[1pt][0pt]}c@{}}{#1}}
\begin{tabularx}{\textwidth}{@{}>{\raggedright\arraybackslash}m{0.235\textwidth} >{\centering\arraybackslash}m{0.085\textwidth} *{6}{>{\centering\arraybackslash}X} >{\centering\arraybackslash}m{0.102\textwidth}@{}}
\toprule
& &
\bcHeader{2}{E6EEF7}{1pt}{1pt}{\textbf{WorldSeek}} &
\bcHeader{2}{E6F3F0}{1pt}{1pt}{\textbf{WorldShape}} &
\bcHeader{2}{F9E6EA}{1pt}{1pt}{\textbf{WorldShift}} &
\bcExecutionHeader{} \\
\multirow{-2}{*}{\textbf{Benchmark}} & \multirow{-2}{*}{\textbf{Setting}} &
\bcHeader{1}{E6EEF7}{1pt}{\tabcolsep}{Nav.} &
\bcHeader{1}{E6EEF7}{\tabcolsep}{1pt}{Ground.} &
\bcHeader{1}{E6F3F0}{1pt}{\tabcolsep}{Geom.} &
\bcHeader{1}{E6F3F0}{\tabcolsep}{1pt}{Phys.} &
\bcHeader{1}{F9E6EA}{1pt}{\tabcolsep}{Memory} &
\bcHeader{1}{F9E6EA}{\tabcolsep}{1pt}{Control} &
\bcExecutionHeader{\multirow{-2}{*}{\textbf{Execution}}} \\
\midrule
\rowcolor[HTML]{F0F1F3}\multicolumn{9}{@{}l@{}}{\rule[-2pt]{0pt}{13pt}\bfseries\itshape Agent and Memory Benchmarks} \\
AgentBench~\cite{liu2023agentbench} & Text / web & \bcP & \bcP & \bcN & \bcN & \bcN & \bcP & \bcP \\[2pt]
ESI-Bench~\cite{closesesi} & Sim. QA & \bcP & \bcY & \bcP & \bcP & \bcP & \bcN & \bcP \\[2pt]
LoCoMo~\cite{maharana2024evaluating} & Dialogue & \bcN & \bcN & \bcN & \bcN & \bcP & \bcN & \bcN \\[2pt]
LongMemEval~\cite{wu2024longmemeval} & Dialogue & \bcN & \bcN & \bcN & \bcN & \bcP & \bcN & \bcN \\[2pt]
MeMoREval~\cite{wang2026should} & Embodied & \bcN & \bcP & \bcN & \bcN & \bcY & \bcN & \bcP \\[2pt]
\rowcolor[HTML]{F0F1F3}\multicolumn{9}{@{}l@{}}{\rule[-2pt]{0pt}{13pt}\bfseries\itshape World Model Benchmarks} \\
Physion~\cite{bear2021physion} & Video & \bcN & \bcN & \bcN & \bcP & \bcN & \bcN & \bcN \\[2pt]
DreamZeroEval~\cite{ye2026world} & Real & \bcN & \bcP & \bcP & \bcN & \bcN & \bcN & \bcY \\[2pt]
\rowcolor[HTML]{F0F1F3}\multicolumn{9}{@{}l@{}}{\rule[-2pt]{0pt}{13pt}\bfseries\itshape Embodied Benchmarks} \\
ALFRED~\cite{shridhar2020alfred} & Sim. & \bcP & \bcP & \bcN & \bcN & \bcN & \bcN & \bcP \\[2pt]
BEHAVIOR-1K~\cite{li2023behavior} & Sim. & \bcN & \bcN & \bcP & \bcP & \bcN & \bcN & \bcY \\[2pt]
LIBERO~\cite{liu2023libero} & Sim. & \bcN & \bcP & \bcP & \bcN & \bcN & \bcN & \bcY \\[2pt]
RoboCasa365~\cite{nasiriany2026robocasa365} & Sim. & \bcP & \bcN & \bcP & \bcN & \bcN & \bcN & \bcY \\[2pt]
RoboDojo~\cite{chen2026robodojo} & Sim. / real & \bcN & \bcP & \bcP & \bcP & \bcP & \bcN & \bcY \\[2pt]
RoboQuest~\cite{renhang2026roboquest} & Sim. & \bcN & \bcY & \bcY & \bcY & \bcN & \bcP & \bcY \\[2pt]
BEHAVIOR-Spatial~\cite{chen2026behavior} & Sim. & \bcP & \bcP & \bcP & \bcN & \bcN & \bcN & \bcY \\[2pt]
IMBench~\cite{maurya2026imbench} & Sim. & \bcN & \bcY & \bcY & \bcY & \bcN & \bcP & \bcY \\[2pt]
RobotAgentBench~\cite{mao2026robotagentbench} & Sim. & \bcP & \bcP & \bcP & \bcN & \bcN & \bcY & \bcY \\[2pt]
Embodied Playground~\cite{chen2026embodied} & Sim. & \bcN & \bcN & \bcY & \bcY & \bcN & \bcP & \bcY \\[2pt]
\specialrule{0.7pt}{4pt}{4pt}
\rowcolor[HTML]{EDF0F6}\textbf{WorldLeague (ours)} & \textbf{Sim.} & \bcY & \bcY & \bcY & \bcY & \bcY & \bcY & \bcY \\
\bottomrule
\end{tabularx}
\endgroup
\end{table}

\section{WorldLeague Benchmark}
\label{sec:benchmark}

\subsection{Physical Goals and Task Demands}
\label{sec:world_challenges}

A human specifies a desired outcome and leaves the agent to decide how to bring it about. To measure the range of goals an embodied AI system can achieve this way, WorldLeague varies what the physical world demands of it. Across 60 tasks, systems explore to find the evidence a goal requires, change the physical configuration of a scene, and adapt as events alter what they must do next. These challenges define the tracks.

\Needspace{3\baselineskip}
\noindent\textbf{WorldSeek.}
Where the robot goes determines what it can reach and what it can see. An ordered pickup tests \emph{spatial navigation}. The robot chooses its route, visits six rooms in the prescribed order, and collects the requested items on its tray. A search for a hidden object tests \emph{goal grounding}. The robot must decide where to look or what to move to expose the candidates, then use what it sees to identify and retrieve the requested object.

\noindent\textbf{WorldShape.}
WorldShape asks the system to rearrange objects in furnished rooms. The room layout determines which objects are within reach, which paths have enough clearance, and which surfaces can support the final arrangement. In a \emph{geometric reconfiguration} task, the robot must orient a cutting board so it clears the sides of a narrow kitchen storage slot. In a \emph{physical reasoning} task, it collects a board and two boxes from separate utility-room counters and chooses their stacking order. The tower must remain stable on the designated surface after release.

\noindent\textbf{WorldShift.}
A change in the scene can determine which object to deliver or force the robot to revisit a completed step. In a \emph{temporal memory} task, the robot must compare its current view with an earlier one, identify the object that changed, and leave it at the destination specified for that change. A kitchen delivery tests \emph{constraint management}. The robot must clear the work area before delivering an apple, then inspect the area again. If the obstruction has returned, the robot must put it back in storage and leave the area clear.

\noindent\begin{minipage}[t]{0.46\linewidth}
\vspace{0pt}
\setlength{\parskip}{0.62\baselineskip}
We assign each task to the dimension that best describes its primary challenge (Table~\ref{tab:task_taxonomy}). Individual tasks draw on several capabilities. Grouping them by their primary demand lets us compare, for instance, a system's success in finding a target with its success in physically reconfiguring one.

Each task specifies what the system is asked to achieve and how the evaluator will recognize it. We write
\end{minipage}\hfill
\begin{minipage}[t]{0.51\linewidth}
\vspace{0pt}
\centering
\footnotesize
\captionsetup{type=table,justification=raggedright}
\caption{WorldLeague groups its 60 tasks by their primary challenge. Counts in parentheses give track totals. Appendix~\ref{sec:task_appendix} lists the task assignments.}
\label{tab:task_taxonomy}
\renewcommand{\arraystretch}{1.05}
\setlength{\tabcolsep}{4.5pt}
\begin{tabular}{lllr}
\toprule
\textbf{Track} & \textbf{Dimension} &
\textbf{Challenge} & \textbf{\#} \\
\midrule

\cellcolor[HTML]{E6EEF7}\textbf{WorldSeek} (22)
& D1~Navigation & Spatial topology & 5 \\
\cellcolor[HTML]{E6EEF7}
& D2~Grounding & Evidence acquisition & 17 \\

\addlinespace[2pt]
\cellcolor[HTML]{E6F3F0}\textbf{WorldShape} (24)
& D3~Geometry & Geometric feasibility & 17 \\
\cellcolor[HTML]{E6F3F0}
& D4~Physics & Action consequences & 7 \\

\addlinespace[2pt]
\cellcolor[HTML]{F9E6EA}\textbf{WorldShift} (14)
& D5~Memory & State evolution & 8 \\
\cellcolor[HTML]{F9E6EA}
& D6~Control & Constrained recovery & 6 \\

\bottomrule
\end{tabular}
\end{minipage}\par

\begin{equation}
    \mathcal{T}_i=(\mathcal{E}_i,\rho_i,g_i,\mathcal{C}_i,\mathcal{G}_i,B_i),
    \label{eq:task_definition}
\end{equation}
where $\mathcal{E}_i$ is the physical scene, $\rho_i$ fixes its initial conditions, $g_i$ is the public goal, and $B_i$ is the interaction budget. The predicates in $\mathcal{G}_i$ specify the requirements of $g_i$, and $\mathcal{C}_i$ specifies the constraints on its pursuit.

\subsection{Task Construction and Validation}
\label{sec:benchmark_quality}

\begin{figure}[H]
\centering
\includegraphics[width=0.94\linewidth]{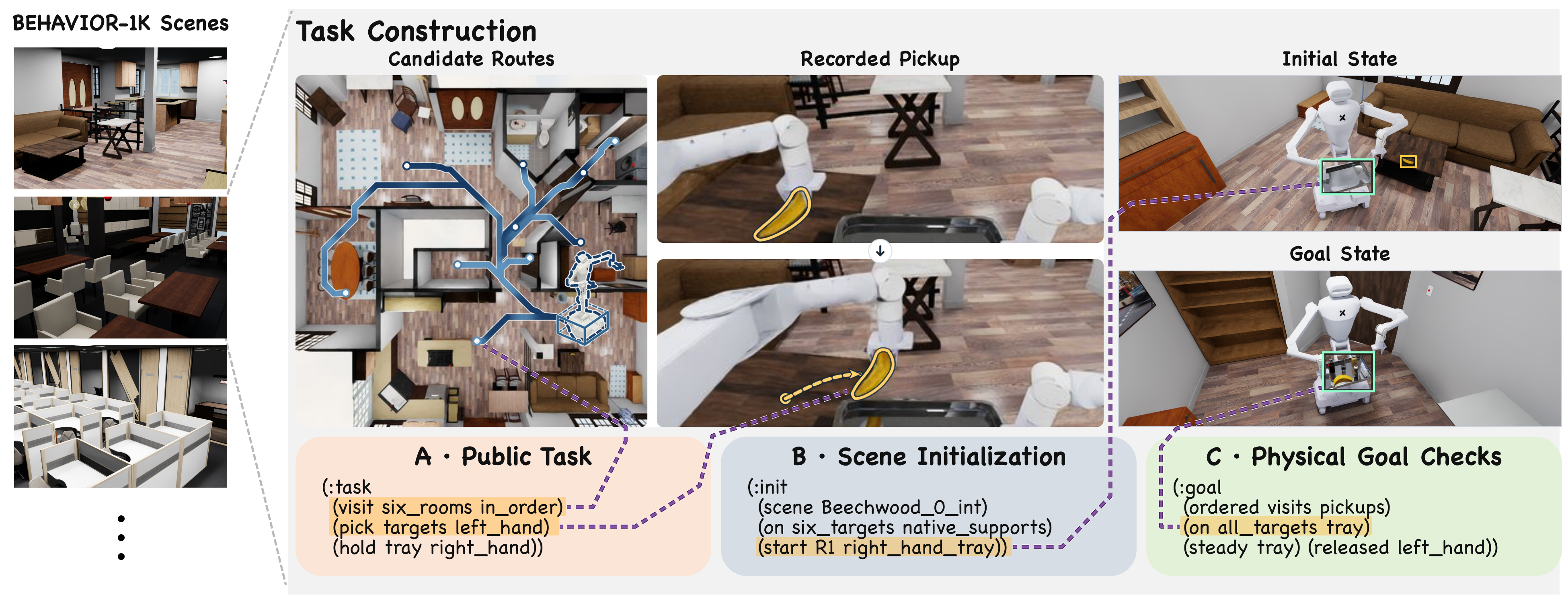}
\caption{From goal to executable task. (A) The instruction specifies the visit order, pickups, and tray constraint. (B) The scene fixes object placements and the robot's starting configuration. (C) The evaluator checks visits and pickups in order, tray contents, stability, and release.}
\label{fig:task_construction}
\end{figure}

For each goal, we write the public instruction, specify the constraints, and define physical predicates for each goal requirement. We bind object roles to scene entities, set their poses and the robot's starting pose, and configure the task events. The evaluator uses the same object bindings to check the private predicates. In the six-room pickup task, the required visit order becomes a sequence of room-entry events, while collection requires the designated objects to remain stable on the tray after release (Fig.~\ref{fig:task_construction}).

We use BEHAVIOR-1K layouts and assets~\cite{li2023behavior} to instantiate the 60 goals, adapting ESI-Bench scene records~\cite{closesesi} for 47 and building scene configurations for the other 13. InteractiveWorld~\cite{wang2026interactiveworld} informed the design of nine challenges. We manually inspected all 60 task--scene pairs, verified that reference runs satisfied their goals and constraints, and checked scoring thresholds against the task specifications (Appendices~\ref{sec:task_construction_provenance} and~\ref{sec:episode_validation}). These fixed instances form the evaluation suite. The generator also instantiates the same goals in new configurations by varying objects, layouts, robot starts, and task events while preserving the scoring predicates (Appendix~\ref{sec:dimension_generation}).

\Needspace{4\baselineskip}
\subsection{Action Interfaces and Decision Responsibility}
\label{sec:harness}

\noindent\begin{minipage}[t]{0.57\linewidth}
\vspace{0pt}
To evaluate how a VLM pursues a goal, we let it choose a target, a motion, and a response when the outcome changes its plan. Human motor control offers a model for this division of responsibility. Sensorimotor feedback coordinates the fine movements needed to carry out a chosen action~\cite{todorov2002optimal,wolpert2011principles}. Show-Harness~\cite{chen2026show} and HumanCLAW~\cite{siyao2026humanclaw} show how an embodied interface can separate action selection from motor execution. In our \emph{decision-preserving action abstraction}, the VLM chooses where to grasp, how far to move, when to release, and what to try next if a motion stops early.

\smallskip
\textbf{WorldHarness} implements this abstraction on R1 by adapting Show-Harness (Fig.~\ref{fig:harness_evaluation}). Given the public goal, camera observations, and retained history, the VLM selects a base, arm, trunk, or gripper command. The controller executes the motion and returns feedback, which the VLM can use to revise its next command. The evaluator records the resulting physical states and task events.

\smallskip
VLA and WAM policies use their native interfaces and trained embodiments to produce control commands or action chunks. All three model families share the task goals, constraints, and physical scoring rules. WorldLeague can therefore compare different policies, harnesses, and robots by the goals each embodied AI system achieves through its own perception, reasoning, and control.
\end{minipage}\hfill
\begin{minipage}[t]{0.40\linewidth}
  \vspace{0pt}\centering
  \includegraphics[width=\linewidth]{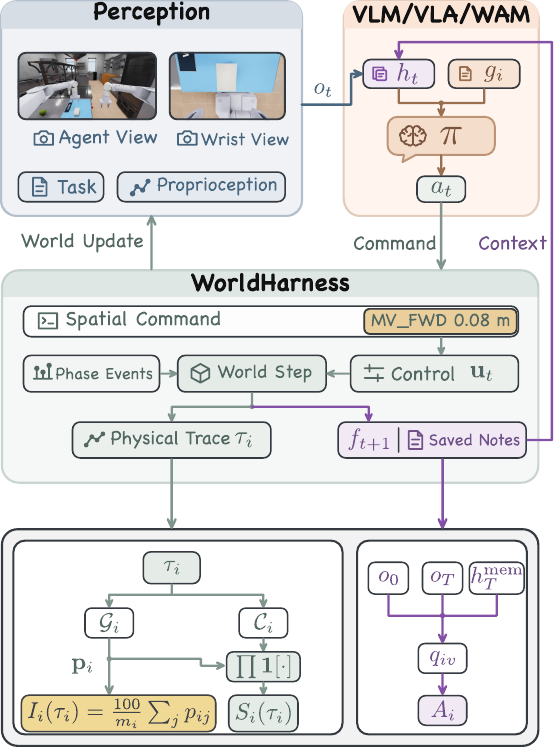}
  \captionsetup{type=figure,font=small,skip=4pt}
  \caption{WorldHarness executes VLM commands and returns observations and feedback. The evaluator scores physical outcomes and terminal QA separately.}
  \label{fig:harness_evaluation}
\end{minipage}\par

\subsection{Measuring Physical Goal Achievement}
\label{sec:physical_evaluation}

A task can require an object to occupy a region at termination, remain stable over an interval, or reach a destination after a specified event. We evaluate these requirements on the episode's timestamped trajectory $\tau_i$ of states, controls, and task events. For each of the $m_i$ goal predicates $G_{ij}$, the evaluator assigns progress $p_{ij}(\tau_i)\in[0,1]$. Each item earns fractional credit where the task defines it and binary credit otherwise. We compute the \emph{Interactive Score} as the mean credit across items, scaled to 100,
\begin{equation}
    I_i(\tau_i)=\frac{100}{m_i}\sum_{j=1}^{m_i}p_{ij}(\tau_i).
    \label{eq:interactive_score}
\end{equation}
A task with an indivisible goal has one item. A required completion declaration is itself a goal predicate. The binary \emph{physical success} indicator checks all goal predicates and constraints together,
\begin{equation}
    S_i(\tau_i)=
    \prod_{j=1}^{m_i}\mathbf{1}[G_{ij}(\tau_i)]\,
    \prod_{C\in\mathcal{C}_i}\mathbf{1}[C(\tau_i)].
    \label{eq:physical_success}
\end{equation}

In T38, the VLM surveys the scene before learning which previously observed objects to deliver. Four completion items record whether it finished the survey, reached the delayed request, selected the requested original objects, and delivered all of them. Completing the first two items alone yields $I_i=50$ and $S_i=0$. These items locate the remaining work at the transition from observing the scene to acting on the remembered identities.

\noindent\textbf{Physical Agency Profiles.}
For each primary dimension $d$, we report mean Interactive Score and physical success rate over its evaluated tasks $\mathcal{T}_d$. Let $\bar I_i$ and $\bar S_i$ be the episode means for task $i$. The pair for $d$ is
\begin{equation}
    (I_d,S_d)=\frac{1}{|\mathcal{T}_d|}\sum_{i\in\mathcal{T}_d}(\bar I_i,\bar S_i).
    \label{eq:dimension_profile}
\end{equation}
These six pairs form the system's \emph{Physical Agency Profile}. Track and overall scores also weight tasks equally. The profile shows where the system makes progress and how often it reaches the full goal.

\noindent\textbf{Terminal question answering.}
For VLMs, we pair the physical outcome with what the model can report about the same episode. The questions appear in the public instruction. After interaction ends, we freeze the physical result and ask the VLM to answer using the initial overview, terminal observations, and retained public memory. The 73 questions cover the initial scene (32), final state (27), and interaction history or changes across phases (14). We match answers to the task's reference values under its answer schema, allowing unordered responses where specified. The percentage correct per task, averaged across tasks, gives a separate measure of scene understanding, outcome recognition, and memory.

\linespread{1.040}\selectfont
\setlength{\parskip}{0.70\baselineskip}
\section{Experiments}
\label{sec:experimental_design}

\subsection{Evaluation Protocol}
\label{sec:experiment_settings}
\label{sec:experiment_configurations}

We give 13 embodied AI systems the same physical goals and score what each achieves through its own perception and control. Eight VLMs direct R1 through WorldHarness. Three VLA and two WAM policies act through their native controllers on PandaOmron or R1Pro. Each VLM runs all 60 tasks. Each continuous-control policy runs 59 tasks, excluding T52, whose containment condition already holds at initialization. With one episode per system and task, this gives 775 episodes at seed~0. We also repeat the full suite with GPT-6.1-Sol at seed~1.

Within each interface family, systems have the same task-specific interaction budgets, ranging from 10 to 342 VLM decisions or 1,000 to 40,000 control steps. Six VLMs select parameterized motions and can revisit earlier images and save notes. Gemini and Jev select a command and a repetition count from current views and public interaction records (Appendix~\ref{sec:analysis_settings}). We report mean Interactive Score ($I$, out of 100) and physical success rate ($S$, in percent), weighting tasks equally. Terminal QA probes the VLM's account of the same episode after its physical outcome is fixed.

\subsection{Physical Goal Achievement Across Systems}
\label{sec:experiment_results}

Claude Fable 5.2 achieves 13 of 60 goals and earns the highest mean Interactive Score, 28.58 (Table~\ref{tab:interactive_results}). Its other 47 episodes split into 12 with partial progress and 35 with zero credit. Across the 59 tasks shared by all systems, Fable achieves 12 goals, Sol eight, and Astra seven. Together the 13 systems achieve 16 goals.

\begin{table}[!htbp]
\centering\footnotesize
\setlength{\tabcolsep}{2.2pt}
\caption{Physical Agency Profiles at seed~0. Each dimension pairs mean Interactive Score ($I$) with the percentage of tasks on which all goals and constraints hold ($S$). Overall covers 60 tasks per VLM and 59 per VLA or WAM. Bold marks each column maximum.}
\label{tab:interactive_results}
\begin{tabular*}{\linewidth}{@{\extracolsep{\fill}}l*{14}{r}@{}}
\toprule
& \multicolumn{4}{c}{\cellcolor[HTML]{E6EEF7}\textbf{WorldSeek}} & \multicolumn{4}{c}{\cellcolor[HTML]{E6F3F0}\textbf{WorldShape}} & \multicolumn{4}{c}{\cellcolor[HTML]{F9E6EA}\textbf{WorldShift}} & \multicolumn{2}{c}{\cellcolor[HTML]{EBEDF3}\textbf{Overall}} \\
& \multicolumn{2}{c}{Navigation} & \multicolumn{2}{c}{Grounding} & \multicolumn{2}{c}{Geometry} & \multicolumn{2}{c}{Physics} & \multicolumn{2}{c}{Memory} & \multicolumn{2}{c}{Constraints} & \multicolumn{2}{c}{All tasks} \\
\cmidrule(lr){2-3}\cmidrule(lr){4-5}\cmidrule(lr){6-7}\cmidrule(lr){8-9}\cmidrule(lr){10-11}\cmidrule(lr){12-13}\cmidrule(lr){14-15}
\textbf{System} & $I$ & $S$ & $I$ & $S$ & $I$ & $S$ & $I$ & $S$ & $I$ & $S$ & $I$ & $S$ & $I$ & $S$ \\
\midrule
\multicolumn{15}{@{}l}{\textit{VLM}} \\
Claude Fable 5.2 & \textbf{46.33} & \textbf{20.00} & \textbf{36.27} & \textbf{29.41} & \textbf{23.53} & \textbf{17.65} & 0.00 & 0.00 & \textbf{45.83} & \textbf{37.50} & 16.67 & 16.67 & \textbf{28.58} & \textbf{21.67} \\
GPT-6.1-Sol & 14.00 & 0.00 & 19.61 & 11.76 & 18.63 & 11.76 & \textbf{14.29} & \textbf{14.29} & 27.08 & 12.50 & \textbf{50.00} & \textbf{50.00} & 22.28 & 15.00 \\
GPT-6-Astra & 17.33 & 0.00 & 19.61 & 11.76 & 15.69 & 11.76 & 0.00 & 0.00 & 27.08 & 12.50 & \textbf{50.00} & \textbf{50.00} & 20.06 & 13.33 \\
DeepSeek-V4.1-Flash~\cite{xu2026deepseek} & 15.67 & 0.00 & 11.76 & 5.88 & 13.73 & 11.76 & 0.00 & 0.00 & 12.50 & 12.50 & 16.67 & 16.67 & 11.86 & 8.33 \\
Kimi-K3.1-Pro~\cite{moonshot2026kimi} & 9.00 & 0.00 & 13.73 & 5.88 & 7.84 & 5.88 & 0.00 & 0.00 & 16.67 & 0.00 & 0.00 & 0.00 & 9.08 & 3.33 \\
Qwen3.8-27B~\cite{qwen38} & 9.00 & 0.00 & 5.88 & 0.00 & 7.84 & 5.88 & 0.00 & 0.00 & 12.50 & 12.50 & 0.00 & 0.00 & 6.31 & 3.33 \\
Gemini-Decision-1 & 9.00 & 0.00 & 8.82 & 0.00 & 7.84 & 5.88 & 0.00 & 0.00 & 0.00 & 0.00 & 0.00 & 0.00 & 5.47 & 1.67 \\
Jev-1.14.0 & 9.00 & 0.00 & 5.88 & 0.00 & 7.84 & 5.88 & 0.00 & 0.00 & 0.00 & 0.00 & 0.00 & 0.00 & 4.64 & 1.67 \\
\midrule
\multicolumn{15}{@{}l}{\textit{VLA}} \\
GR00T-N1.5~\cite{gr00tn1} & 10.33 & 0.00 & 17.65 & 11.76 & 4.17 & 0.00 & 0.00 & 0.00 & 12.50 & 12.50 & 0.00 & 0.00 & 8.79 & 5.08 \\
$\pi_{0.5}$~\cite{intelligence2025pi} (R1Pro SFT) & 9.00 & 0.00 & 15.69 & 11.76 & 2.08 & 0.00 & 0.00 & 0.00 & 16.67 & 12.50 & 0.00 & 0.00 & 8.11 & 5.08 \\
Xiaomi XR-1~\cite{team2026xiaomi} & 0.00 & 0.00 & 5.88 & 0.00 & 4.17 & 0.00 & 0.00 & 0.00 & 0.00 & 0.00 & 0.00 & 0.00 & 2.82 & 0.00 \\
\midrule
\multicolumn{15}{@{}l}{\textit{WAM}} \\
Awomo-0.5I~\cite{team2026awomo} & 13.00 & 0.00 & 11.76 & 5.88 & 2.08 & 0.00 & 0.00 & 0.00 & 12.50 & 12.50 & 0.00 & 0.00 & 6.75 & 3.39 \\
OpenWAM-Alpha~\cite{wang2026openwam} & 11.67 & 0.00 & 5.88 & 0.00 & 4.17 & 0.00 & 0.00 & 0.00 & 0.00 & 0.00 & 0.00 & 0.00 & 3.81 & 0.00 \\
\bottomrule
\end{tabular*}
\end{table}

Fable leads in navigation, goal grounding, geometric reconfiguration, and temporal memory on both measures. Sol and Astra share the lead in constraint management. Both complete three of the six tasks, with two successes in common. Astra clears a returning obstruction on a delivery where Sol scores zero. Sol completes an object-allocation task on which Astra scores zero. Sol also builds the three-object tower in T21, the only physical-reasoning task any system completes at seed~0.

Among continuous-control policies, GR00T and $\pi_{0.5}$ each complete three goals, and Awomo completes two. All three succeed on T10, where the robot must bring both a cabinet and a kitchen table into view, and on T41, where it must compare the contents of boxes across two phases. GR00T and $\pi_{0.5}$ also succeed on T09. Four of the five policies earn navigation credit, and all five earn geometric reconfiguration credit. None completes a navigation or geometric reconfiguration task. On T05, for example, Awomo earns 40 points by satisfying the requirements in two of five rooms. GR00T satisfies the requirement in one room and earns fractional credit in a second.

\vspace{6pt}
\subsection{Task Coverage and Unmet Goal Requirements}
\label{sec:experiment_coverage}

Across the 59 common tasks, every system scores zero on 27, at least one system makes partial progress but none succeeds on 16, and at least one achieves the goal on 16 (Table~\ref{tab:shared_task_coverage} and Appendix Fig.~\ref{fig:task_outcomes}). At least one system makes progress on each of the five navigation tasks. Only one of the five is completed. In WorldShape, all systems score zero on 15 of its 23 common tasks, including six of seven physical-reasoning tasks. In constraint management, at least one system completes each of the four tasks that earn any credit.

Even full item credit can leave a constraint unmet. On T41, Kimi earns $I=100$ for comparing the scene across both phases. A separate constraint fails, so the episode ends in physical failure (Appendix~\ref{sec:episode_success}).

\begin{table}[!htbp]
\centering\small
\setlength{\tabcolsep}{5pt}
\caption{Coverage of the 59 common tasks at seed~0. For each task we check all 13 systems. A task counts as partial if some system earns progress and none succeeds.}
\label{tab:shared_task_coverage}
\begin{tabular}{@{}lrrrrrrr@{}}
\toprule
Best outcome across systems & Navigation & Grounding & Geometry & Physics & Memory & Constraints & Total \\
\midrule
Zero progress & 0 & 8 & 9 & 6 & 2 & 2 & 27 \\
Partial progress & 4 & 4 & 5 & 0 & 3 & 0 & 16 \\
Physical success & 1 & 5 & 2 & 1 & 3 & 4 & 16 \\
\bottomrule
\end{tabular}
\end{table}

\begin{figure}[!htbp]
    \centering
    \includegraphics[width=\linewidth]{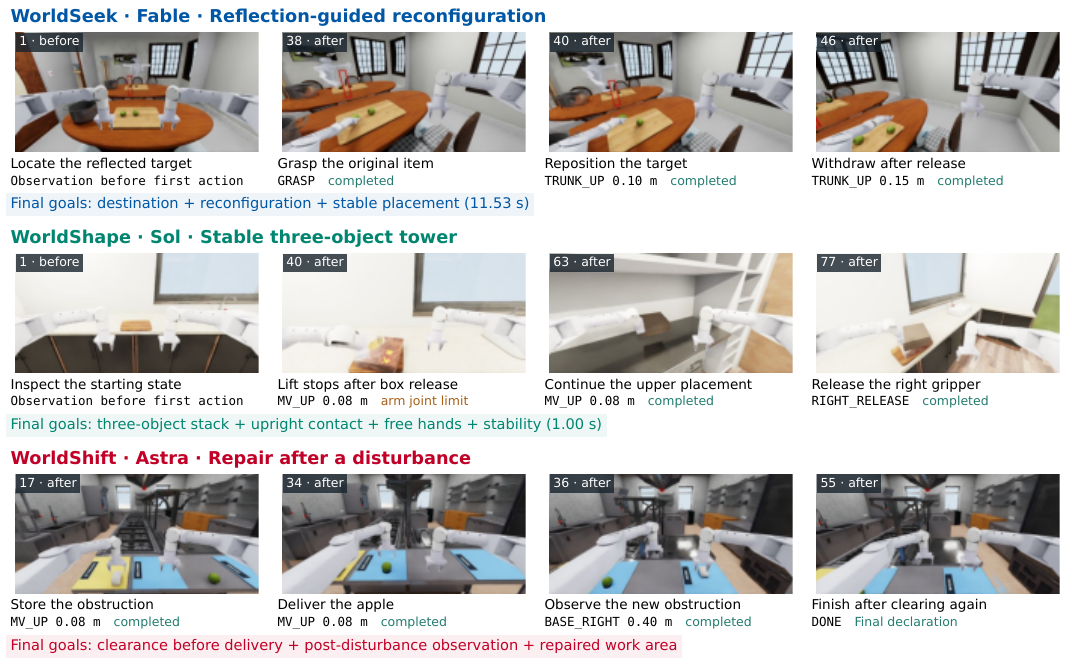}
    \caption{Three goals achieved through interaction. Fable uses a reflection to reposition an object (top), Sol builds a stable tower after an arm joint-limit stop (middle), and Astra clears an obstruction that returns after delivery (bottom). Frame labels identify the decision and whether the view precedes or follows it.}
    \label{fig:physical_trajectories}
\end{figure}

\noindent\textbf{Reaching a destination and leaving the object in place.}
ESI-Bench~\cite{closesesi} reports errors in matching mirror reflections to their corresponding objects. Across WorldLeague's three mirror-guided rearrangements, the selected object reaches its destination in 24 of 24 VLM episodes, eight of nine VLA episodes, and all six WAM episodes. Only Fable on R-S1 also establishes the required spatial relation and leaves the object stable after release (Fig.~\ref{fig:physical_trajectories}, top). It grasps the object, raises the robot's trunk, releases the object, and withdraws. The object remains stable for 11.53 seconds. On T32, all eight VLMs achieve the requested separation between objects, yet none places an object on the tray.

\noindent\textbf{Using earlier evidence to choose a later action.}
On T38, the robot inspects nine food items before it learns which ones to deliver. Fable, Sol, and Astra all finish the survey and reach the delayed request. Only Fable delivers the requested original objects. Across five change-guided delivery tasks, eight of 40 VLM episodes complete the two-phase comparison. One of those eight, Fable on T46, identifies the changed object and leaves it stably at the destination specified for that change.

\subsection{Decisions, Feedback, and Termination}
\label{sec:experiment_decisions}

WorldHarness records 32,455 decisions across 480 VLM episodes. For Qwen, 3,127 of 4,440 decisions yield no valid command. Jev completes 95.26\% of the motions it initiates, yet succeeds only on T52, where the goal already holds at the start. Sol completes 70.74\% of its motion commands and achieves nine goals.

\noindent\textbf{Which decisions does the interface leave to the agent?}
With object-level grasp and placement skills, Sol must still bring the target within reach. Of its 52 named grasp attempts, 30 target objects outside the workspace. On T24, Sol completes the task with motion commands and scores zero with the skills available.

Even after a system resumes moving, the original obstacle can persist. Across 6,241 three-decision windows following a stopped command, 88.75\% contain a completed command and 56.55\% include a recurrence of the original stop status. Both occur in 47.54\% of windows (Appendix~\ref{sec:decision_statistics}). In T21, Sol encounters an arm joint limit after placing the first box. It goes on to place the second box and release both hands, leaving an upright, stationary tower. In the kitchen delivery, Astra clears the work area, delivers the apple, then sees that the obstruction has returned. It puts the obstruction back in storage before declaring completion (Fig.~\ref{fig:physical_trajectories} and Appendix~\ref{sec:trajectory_evidence}).

On T22, Sol completes placement with full execution feedback and declares completion at decision~41. With status, failure reasons, and recovery hints hidden, placement is still unverified at the 52-decision limit. Both runs answer the terminal question correctly.

The agent also chooses when it has done enough. Of 279 episodes that end with a final \texttt{DONE}, 41 satisfy the physical goal, 42 earn credit with a goal or constraint unmet, and 196 earn zero. Of the 201 episodes that exhaust the budget, 31 earn partial credit and 170 earn zero (Appendix~\ref{sec:episode_termination}). On the allocation task, Sol declares completion at decision~74 with saved notes and uses all 94 decisions with notes hidden. Both runs satisfy three of four placement checks and miss different placements (Appendix~\ref{sec:sol_diagnostics}).

\subsection{What VLMs Can Report About an Episode}
\label{sec:experiment_analysis}

\begin{figure}[!htbp]
    \centering
    \includegraphics[width=\linewidth]{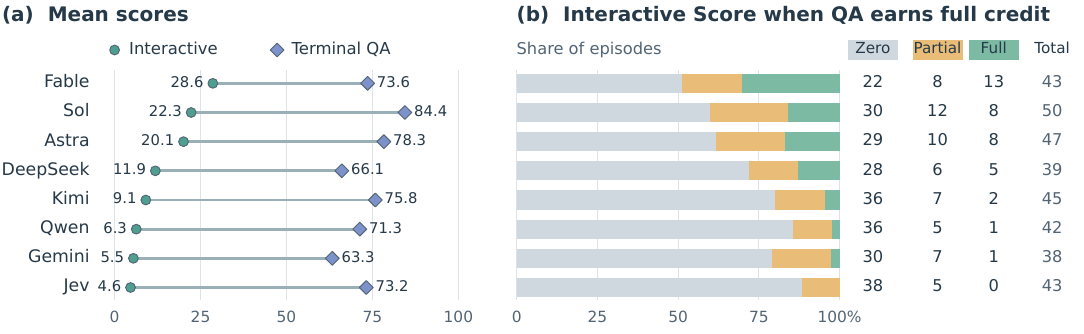}
    \caption{What eight VLMs report and achieve in the same episode. Left, mean Interactive Score and terminal QA over all tasks. Right, the distribution of zero, partial, and full Interactive Scores among the episodes in which each VLM answers every question correctly.}
    \label{fig:qa_execution}
    \vspace{12pt}
\end{figure}

A VLM can correctly describe an unsuccessful attempt. Of 347 VLM episodes with full terminal QA credit, 249 (71.8\%) earn zero Interactive Score. The zero-credit rate across all 480 VLM episodes is 76.3\%. Figure~\ref{fig:qa_execution} shows this pairing for each VLM. Appendix~\ref{sec:qa_episode_types} reports results across initial-state, final-state, and history questions. On T42, for example, Sol correctly reports that no object completed a valid physical trial. T21 shows the converse. Sol builds a stable tower and answers the question about the maximum-stability ordering incorrectly.

On 12 tasks, we also ask Sol the initial-state question before its first action. We keep its answer out of the control context and ask the same question again at termination. Sol answers six correctly at the start and ten at the end. Of the six it answers correctly at the start, it completes three physical goals and earns zero progress on the other three (Appendix~\ref{sec:sol_diagnostics}).

\vspace{12pt}
\subsection{Variation Across Runs and Computational Cost}
\label{sec:experiment_resources}

Sol's overall Interactive Score changes from 22.28 to 21.44 across two environment seeds, yet its score changes on 11 of 60 tasks. Sol builds the T21 tower only at seed~0 and completes the T46 change-guided delivery only at seed~1. These gains and losses nearly cancel in the overall score (Appendix~\ref{sec:sol_seed_repeat}).

\begin{figure}[!htbp]
    \centering
    \includegraphics[width=\linewidth]{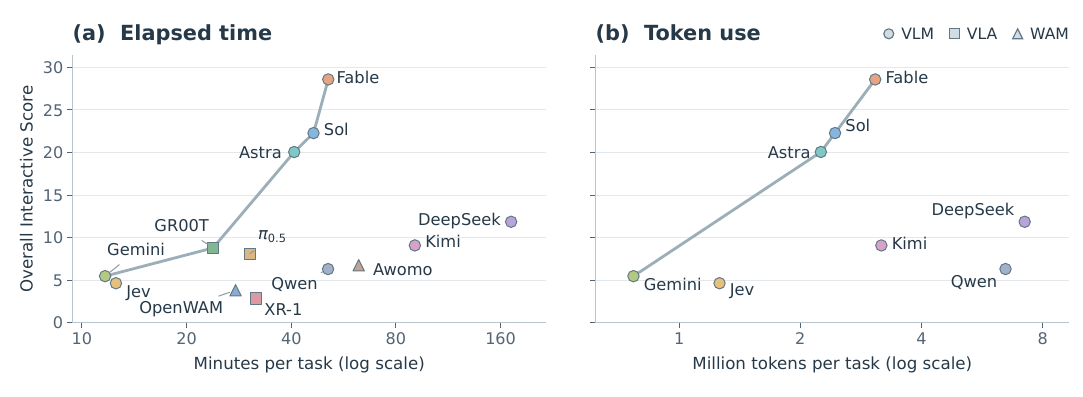}
    \caption{Overall Interactive Score versus mean elapsed minutes per task (left, all models) and mean millions of tokens per task (right, VLMs). Both horizontal axes are logarithmic. Lines connect the observed Pareto-optimal points. Colors identify models and shapes distinguish model families.}
    \label{fig:score_efficiency}
\end{figure}

\vspace{12pt}
Figure~\ref{fig:score_efficiency} compares physical progress with mean elapsed time for all 13 systems and mean token use for the eight VLMs. Gemini-Decision is at the low-resource end of both observed Pareto frontiers, and GR00T is the only continuous-control policy on the elapsed-time frontier. Fable and Qwen each take about 51 minutes per task. Fable earns the highest score, 28.58, with 3.07 million tokens per task, and Qwen earns 6.31 with 6.47 million.

GR00T spends 16.39 of its 23.83 minutes per task in simulator execution and 1.08 in inference. Sol spends 14.74 of its 46.36 minutes in inference. Total elapsed time also includes observation and harness overhead (Appendix~\ref{sec:resource_appendix}). Across the eight VLMs, terminal QA accounts for 0.10--2.70\% of inference time, so most model-request time is spent choosing actions during physical interaction.

\vspace{4pt}
As in the physical world, the environment keeps running during model inference (Appendix~\ref{sec:analysis_settings}). The scene can therefore change between observation and action execution, making inference latency a practical consideration for physical interaction.

\section{Conclusion}
\label{sec:conclusion}

{\setlength{\parfillskip}{0pt}%
WorldLeague evaluates physical agency across 60 tasks in exploration, physical transformation, and adaptation. Across 13 models, the highest Interactive Score is 28.58 out of 100. Every VLM scores higher on auxiliary QA than on physical operation, and completed commands often leave physical goals unmet. Of the 347 VLM episodes that receive full credit for terminal QA, 249 (71.8\%) receive zero operation credit for the physical task. WorldHarness traces these outcomes to the agent's choice of targets and motion parameters and to its subsequent actions. The same predicates that determine the capability profile show which requirements remain unmet in each trajectory, from stable placement after release to finding and delivering the right object after a scene change. The world continues to evolve while a model reasons, making inference latency relevant to the evaluation of physical interaction.\par}
\noindent\textbf{Future Work.}
We could adapt JEPA-based planners such as V-JEPA 2-AC~\cite{assran2025v} to WorldLeague to test whether action-conditioned latent predictions help with geometric reconfiguration and recovery from disturbances. For these experiments, we would use the generator in Appendix~\ref{sec:dimension_generation} to sample a template $i$ with fixed probability $w_i$ and then draw an instance $z$ from $p_i$, yielding $p(z,i)=w_i p_i(z)$. We would group instances by scene, object-role assignment, layout, and events before assigning groups to splits, retaining the 60 fixed evaluation instances as the test set.
The same generator could supply a screened pool for online reinforcement learning. At each reset, the environment would sample an instance from the pool, initialize the objects and robot pose, clear the episode history, and install the task events. The policy would receive the public goal, observations, and feedback through the interaction interface (Sec.~\ref{sec:harness}), and physical progress and terminal success could serve as reward signals.
\linespread{1}\selectfont

{
    \small
    \setlength{\parfillskip}{0pt plus 1fil}
    \bibliographystyle{ieeenat_fullname_shuffled}
    \bibliography{main}
}

\clearpage
\raggedbottom
\setlength{\parskip}{0.55\baselineskip}
\setlength{\intextsep}{10pt plus 2pt minus 2pt}
\setlength{\textfloatsep}{16pt plus 2pt minus 3pt}
\appendix
\section{Physical Goals and Action Interfaces Across Benchmarks}
\label{sec:benchmark_comparison_appendix}

\subsection{What Counts as Success?}
\label{sec:closest_benchmark_protocols}

A robot can gather evidence, act on it, and check the outcome within one episode. Where a benchmark measures success determines which of those decisions contribute to the score. In RoboQuest the robot makes trial impressions before submitting a stamped pattern. IMBench grades understanding and planning in separate stages before execution. RobotAgentBench follows a household activity through disturbances and recovery. Table~\ref{tab:closest_benchmark_protocols} compares their observation, control, and scoring protocols with WorldLeague's.

\begin{table}[!htbp]
\centering\small
\setlength{\tabcolsep}{4pt}
\renewcommand{\arraystretch}{1.12}
\caption{How four benchmarks connect evidence, robot control, and task completion.}
\label{tab:closest_benchmark_protocols}
\begin{tabularx}{\linewidth}{@{}>{\raggedright\arraybackslash}p{0.18\linewidth}*{3}{>{\raggedright\arraybackslash}X}@{}}
\toprule
Benchmark & What the agent observes & What the agent controls & What earns credit \\
\midrule
RoboQuest~\cite{renhang2026roboquest}
& Cameras and proprioception reveal hidden properties through inspection and physical trials.
& Timed arm, base, and gripper commands. The robot presses a physical submit button.
& Goal conditions at submission, such as placing marked cubes in their assigned bins. Trace rules identify missed evidence, decisions, execution, and side effects. \\
\addlinespace
IMBench~\cite{maurya2026imbench}
& Stage~3 supplies images and robot signals. A privileged variant also supplies object poses.
& Target end-effector poses, gripper primitives, or raw controls guide interventions such as intercepting a moving object.
& Understanding, planning, and execution receive separate scores. Physical checks include insertion depth and alignment after perturbation. \\
\addlinespace
RobotAgentBench \cite{mao2026robotagentbench}
& Cameras, proprioception, and a progressively revealed map. Demonstration generation also uses simulator state.
& The C/C/C agent composes motion functions. Other configurations delegate to learned policies.
& Target contact, progress, and recovery during activities such as replacing a tablecloth. Disturbance variants measure recovery and stagnation. \\
\addlinespace
WorldLeague
& Public views and execution feedback, with history tools by configuration. The evaluator observes physical states and events.
& VLMs direct R1 through WorldHarness. VLA and WAM policies use native control.
& Individual goal predicates earn progress. Satisfying the full goal and its constraints earns physical success. VLMs then answer questions about the same episode. \\
\bottomrule
\end{tabularx}
\end{table}

\subsection{Who Chooses the Next Motion?}
\label{sec:harness_adaptation}

Show-Harness translates increments of end-effector motion, rotation, and gripper control into robot commands. Its plugins provide multiple views, proprioception, action history, adaptive increments, replanning, and automatic grasp recovery. It measures task success and steps per episode~\cite{chen2026show}. WorldHarness adapts that interface to the mobile R1 in OmniGibson. The VLM can reposition the base, raise the trunk, translate or rotate the wrist, and grasp or release. It chooses the next command after seeing how far the robot moved and why a motion stopped. Six VLMs set motion parameters directly. Gemini-Decision-1 and Jev-1.14.0 select a command and repetition count.

\noindent\textbf{A shared scene can support different goals.}\label{sec:esi_harness_comparison}
WorldLeague adapts selected ESI-Bench scenes built on BEHAVIOR-1K~\cite{closesesi,li2023behavior}. ESI-Bench uses interaction to reveal evidence for a spatial answer. In WorldLeague's R-S1 scene, Sol encounters a stop on a 0.25\,m base advance and completes a subsequent 0.08\,m arm advance. At termination the object is at its destination, but its relation and stability checks fail. Sol earns 33.33 for physical progress and 100 on terminal QA. Table~\ref{tab:esi_harness_comparison} shows how the action interface and endpoint differ.

\begin{table}[H]
\centering\small
\renewcommand{\arraystretch}{1.15}
\caption{System interfaces and scored outcomes in ESI-Bench~\cite{closesesi} and WorldLeague on BEHAVIOR-1K.}
\label{tab:esi_harness_comparison}
\begin{tabularx}{\textwidth}{@{}p{0.19\textwidth}XX@{}}
\toprule
 & \textbf{ESI-Bench} & \textbf{WorldLeague with WorldHarness} \\
\midrule
Action interface & Movement, viewpoint changes, and object-level pickup, placement, filling, and pouring & Parameterized base and end-effector movement, wrist rotation, trunk movement, and grasp/release \\
Agent choices & What to inspect or manipulate and in which order to acquire evidence & Which movements and parameters to compose, and how to revise them from feedback \\
Execution support & Realization of selected high-level interactions & Motor execution of selected robot commands \\
Scored outcome & Answer after exploration & Goal and constraint satisfaction on the trajectory, followed by separate QA \\
\bottomrule
\end{tabularx}
\end{table}

\subsection{Reading the Six Capability Dimensions}
\label{sec:comparison_criteria}

Table~\ref{tab:benchmark_comparison} marks a dimension as fully covered when the benchmark scores its demands together. Partial coverage means that an explicit test measures one component. The Execution column records who controls the motions. Feedback-guided robot motion earns a full mark, delegated skills, symbolic commands, and exploration for QA earn a partial mark, and passive prediction has no execution mark.

\noindent\textbf{WorldSeek.} Navigation links route or region choices to physical arrival. Arrival checks, symbolic traversal, and route questions each test a component. Goal grounding links identification of the intended target to action that reveals decisive evidence~\cite{huang2026era}. Selecting a visible referent tests the identification component.

\noindent\textbf{WorldShape.} Geometric reconfiguration couples choices about shape, clearance, or orientation to an achieved arrangement. Relational placement and insertion alignment test individual components. Physical reasoning links a prediction about an action's consequences to an intervention, such as stabilizing a support or choosing a launch. Contact predictions and material-outcome checks each test a component.

\noindent\textbf{WorldShift.} Temporal memory requires the agent to use earlier physical evidence when a changed scene calls for a decision. Dialogue recall and reproduction of a demonstrated sequence each test memory. Constraint management couples task rules or resource limits with recovery under disturbance. Enforcing a rule or testing recovery in isolation covers one component.

\subsection{Tasks Behind the Comparison}

\noindent\textbf{Exploration and memory.}
AgentBench~\cite{liu2023agentbench} tests symbolic household traversal and target selection in ALFWorld~\cite{shridhar2020alfworld}, and decisions under enforced rules in card games. ESI-Bench~\cite{closesesi} asks agents to uncover occluded targets and answer questions about routes, arrangements, physical outcomes, and earlier states. LoCoMo~\cite{maharana2024evaluating} tests temporal and multi-hop recall over conversations. LongMemEval~\cite{wu2024longmemeval} tests recall across sessions, temporal reasoning, and knowledge updates. MeMoREval~\cite{wang2026should} carries earlier evidence into a physical request. In \texttt{spatial\_shelf\_vacate}, a figurine's former shelf location identifies the book to retrieve, and the agent uses complete skills to execute its plan.

\noindent\textbf{Prediction and action.}
Physion~\cite{bear2021physion} tests whether a model can predict object contact from the opening frames. CoPhyBench tests physical reasoning from an observed video segment~\cite{weicophybench}. Video world models can use spatial memory to predict future frames more efficiently~\cite{shen2026velox}. World-action models use predictions of future states to guide action selection~\cite{shen2026world}. Under visual perturbations, a model can predict a plausible outcome yet execute an action that fails~\cite{li2026badwam}. DreamZero~\cite{ye2026world} executes action chunks on AgiBot and DROID and observes the result before acting again. Its tasks include object matching, orientation, and shape-matched insertion, with partial credit on DROID.

\noindent\textbf{Household activities.}
ALFRED~\cite{shridhar2020alfred} combines egocentric navigation and target masks with simulator actions for pickup, placement, and object-state changes. BEHAVIOR-1K~\cite{li2023behavior} scores spatial, cloth, and material states through BDDL goals~\cite{srivastava2022behavior}. RL-VMC issues joint commands and RL-Prim selects skills. In LIBERO~\cite{liu2023libero}, spatial descriptions identify the target and \texttt{On}/\texttt{In} predicates score placement. RoboCasa365~\cite{nasiriany2026robocasa365} checks base position and heading in navigation, size in nesting, and pan tilt and final containment in \texttt{TiltPan}.

\noindent\textbf{Manipulation guided by evidence.}
RoboDojo~\cite{chen2026robodojo} scores target identity, insertion depth and alignment, transferred fluid volume, and reproduction of a sorting sequence. In RoboQuest~\cite{renhang2026roboquest}, \texttt{Stamps} scores the pattern produced by contact and \texttt{WobblyStand} requires selecting shims that support a ball. Hidden targets reward inspection and protected objects constrain manipulation. BEHAVIOR-Spatial~\cite{chen2026behavior} checks arrival near a target, referent selection, and spatial relations such as the side of a knife or the height of a shelf placement.

\noindent\textbf{Dynamics and recovery.}
IMBench~\cite{maurya2026imbench} pairs occluder removal, insertion, and balancing with forced-slip and stack-collapse tests. Its Stage-3 VLM uses \texttt{MoveEEF} and gripper commands with updated observations. RobotAgentBench~\cite{mao2026robotagentbench} scores navigation, contextual target selection, placement, and cloth coverage under contact budgets and interference. Its C/C/C agent composes motion functions and checks their outcomes. The C/L/-- planner delegates subtasks to a learned executor. Embodied Playground~\cite{chen2026embodied} checks puzzle configurations, target impacts, legal moves, ammunition, and target order in manipulation and launch tasks.

\FloatBarrier
\section{Tasks and Evaluation Scenes}
\label{sec:task_appendix}

WorldSeek asks the robot to acquire evidence, WorldShape asks it to change a configuration, and WorldShift makes its next action depend on an earlier observation or a disturbance. Table~\ref{tab:task_assignments} locates each task by its main demand. The same task ID links the goal, its scene, and the outcomes for all 13 systems.

\begin{table}[!htbp]
\centering\small
\caption{WorldLeague's 60 tasks grouped by their primary demand. Each task ID labels a column in Fig.~\ref{fig:task_outcomes}.}
\label{tab:task_assignments}
\setlength{\tabcolsep}{4pt}
\renewcommand{\arraystretch}{1.0}
\begin{tabularx}{\textwidth}{l l >{\raggedright\arraybackslash}X r}
\toprule
\textbf{Track} & \textbf{Primary demand} & \textbf{Task IDs} & \textbf{Count} \\
\midrule
\rowcolor[HTML]{E6EEF7}
\textbf{WorldSeek} & Navigation (D1) & T04--T08 & 5 \\
& Goal grounding (D2) & T09--T11, T35, T36, T49, T50; R-C1--R-C3, R-S1--R-S3, R-A1, R-A2; \texttt{iw\_h1\_02}, \texttt{iw\_m3\_01} & 17 \\
\addlinespace[5pt]
\rowcolor[HTML]{E6F3F0}
\textbf{WorldShape} & Geometry (D3) & T15--T17, T22--T26, T28, T30, T32, T33, T52; GF2-1, GF2-2, GF3-1, GF3-2 & 17 \\
& Physical reasoning (D4) & T01, T02, T18, T21, T42--T44 & 7 \\
\addlinespace[5pt]
\rowcolor[HTML]{F9E6EA}
\textbf{WorldShift} & Temporal memory (D5) & T38, T40, T41, T45--T48; \texttt{iw\_m1\_01} & 8 \\
& Constraint management (D6) & \texttt{iw\_h2\_02}, \texttt{iw\_h3\_03}, \texttt{iw\_h4\_02}, \texttt{iw\_h5\_02}, \texttt{iw\_m2\_01}, \texttt{iw\_m4\_02} & 6 \\
\bottomrule
\end{tabularx}
\end{table}

\subsection{From a Goal to an Executable Scene}
\label{sec:task_construction_provenance}

We specify the public instruction, object roles, constraints, and physical predicates before placing the robot in the scene. We then bind each role to an object, set its pose and the robot's starting pose, and specify any events that occur during interaction. We inspected all 60 task--scene pairs and verified in reference runs that the goals and constraints could be satisfied. Appendix~\ref{sec:episode_validation} gives the physical checks, and Appendix~\ref{sec:dimension_generation} describes how the same goals can be instantiated in other configurations.

The fixed suite spans 23 scene identifiers (Table~\ref{tab:task_sources_1}). We adapted ESI-Bench scene records~\cite{closesesi} for 47 instances, including T52, whose containment goal holds at initialization. For the other 13, we built scenes for nine \texttt{iw} tasks and four household geometry tasks. The latter use a carton cubby, a cutting-board slot, cucumber compartments, and spice-jar shelves to vary orientation and clearance. One layout can support distinct goals. R-A1 and R-A2 share a base scene for direct and reflection-guided tasks.

\begin{table}[!htb]
\centering\footnotesize
\setlength{\tabcolsep}{4pt}
\renewcommand{\arraystretch}{0.97}
\caption{WorldLeague's 60 tasks indexed by scene and room. E marks an adapted ESI-Bench scene record and W a scene configuration built for WorldLeague. Table~\ref{tab:task_assignments} gives the primary demand of each task.}
\label{tab:task_sources_1}
\begin{tabularx}{\linewidth}{@{}l l >{\raggedright\arraybackslash}X c@{}}
\toprule
Scene & Room & Task IDs & Source \\
\midrule
Beechwood\_0\_garden & dining\_room\_0 & T15, T18, T41, T45, R-C1, R-S1 & E \\
 & kitchen\_0 & T01, T26, T28 & E \\
Beechwood\_0\_int & full\_scene & T07 & E \\
 & living\_room\_1 & T22 & E \\
Beechwood\_1\_int & television\_room\_0 & T23 & E \\
grocery\_store\_asian & grocery\_store\_0 & iw\_m2\_01 & W \\
grocery\_store\_cafe & grocery\_store\_0 & iw\_h5\_02 & W \\
hall\_arch\_wood & empty\_room\_0 & T43, T44, T49, T50 & E \\
hotel\_suite\_large & bedroom\_0 & iw\_m1\_01 & W \\
house\_double\_floor\_upper & full\_scene & T05 & E \\
house\_single\_floor & dining\_room\_0 & T02, T40, R-C3, R-S3 & E \\
 & living\_room\_0 & T38, T42 & E \\
 & utility\_room\_0 & T21, T24, T30, T33 & E \\
Ihlen\_0\_int & living\_room\_0 & T36 & E \\
Merom\_0\_garden & playroom\_0 & T25, T32 & E \\
Merom\_0\_int & playroom\_0 & T35 & E \\
office\_cubicles\_left & full\_scene & T08 & E \\
 & meeting\_room\_0 & T47, T48, R-A1, R-A2, R-C2, R-S2 & E \\
office\_large & meeting\_room\_1 & T52 & E \\
restaurant\_asian & kitchen\_0 & iw\_h3\_03, iw\_m3\_01 & W \\
restaurant\_cafeteria & full\_scene & T09, T10 & E \\
restaurant\_diner & kitchen\_0 & iw\_m4\_02 & W \\
restaurant\_hotel & kitchen\_0 & GF2-1, GF2-2, GF3-1, GF3-2, iw\_h1\_02, iw\_h2\_02 & W \\
 & lobby\_0 & T16 & E \\
Rs\_int & full\_scene & T04 & E \\
school\_chemistry & chemistry\_lab\_0 & iw\_h4\_02 & W \\
school\_geography & full\_scene & T06 & E \\
school\_gym & full\_scene & T11 & E \\
 & locker\_room\_1 & T17 & E \\
Wainscott\_0\_int & dining\_room\_0 & T46 & E \\
\bottomrule
\end{tabularx}
\end{table}

\FloatBarrier
\section{Task Outcomes and Evaluation Settings}
\label{sec:task_outcome_analysis}

\begin{figure}[!htb]
    \centering
    \includegraphics[width=\textwidth]{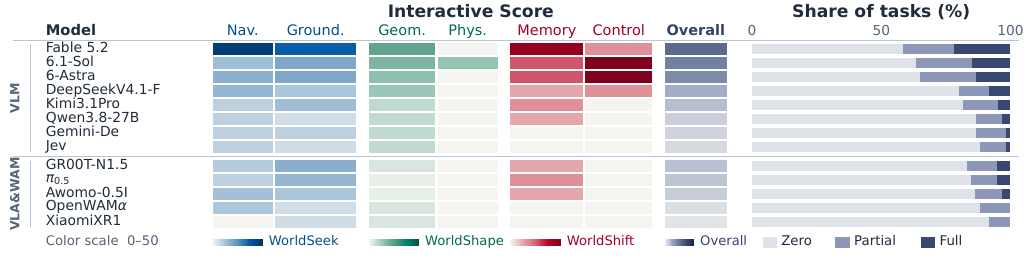}
    \caption{The left panel shows mean Interactive Scores by dimension and overall (darker is higher, color scale 0--50). The right panel shows each model's share of evaluated tasks with zero ($I=0$), partial ($0<I<100$), and full ($I=100$) credit.}
    \label{fig:capability_distribution}
\end{figure}

\begin{figure}[!htb]
\centering
\includegraphics[width=0.97\textwidth,height=0.69\textheight,keepaspectratio]{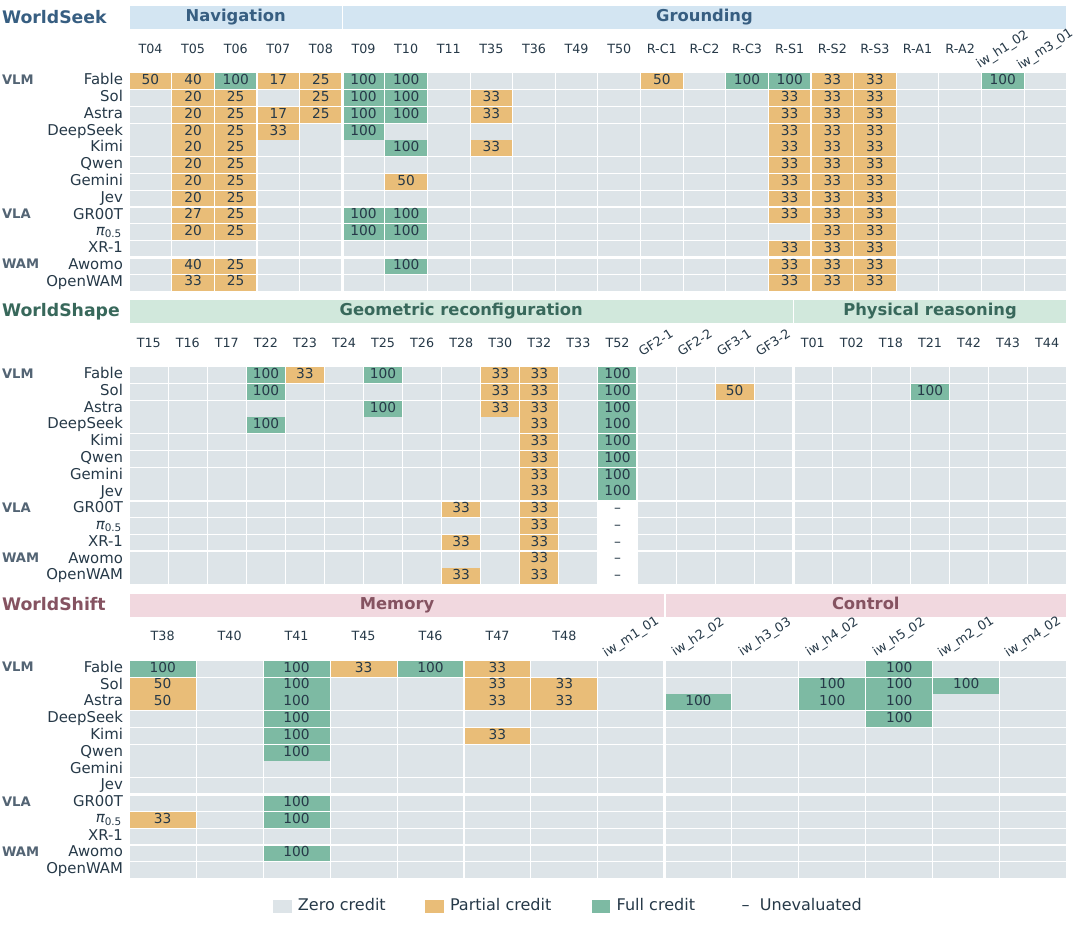}
\caption{Task-level Interactive Scores by model. Each panel groups tasks from one track by dimension. Gray, amber, and green mark zero, partial, and full credit. Numbers show nonzero scores rounded to integers. Dashes mark the unevaluated T52 task for continuous-control policies.}
\label{fig:task_outcomes}
\end{figure}

On the 59 shared goals, the highest mean Interactive Score is 27.37. On the full suite, Fable earns partial credit on 12 tasks and full credit on 13. Sol earns partial credit on 13 and full credit on nine. Table~\ref{tab:task_outcome_counts} gives both denominators and the outcome counts for every system. Figure~\ref{fig:capability_distribution} places these outcomes alongside each system's six-dimensional profile. Figure~\ref{fig:task_outcomes} identifies the individual tasks behind the averages.

\begin{table}[!htb]
\centering\small
\setlength{\tabcolsep}{7pt}
\caption{Mean Interactive Score over all evaluated tasks and the common 59 excluding T52. The last three columns count tasks by Interactive Score over all $N$ tasks.}
\label{tab:task_outcome_counts}
\begin{tabular}{@{}lrrrrrr@{}}
\toprule
\textbf{Model} & $N$ & \textbf{All tasks} & \textbf{Common 59} & \textbf{Zero} & \textbf{Partial} & \textbf{Full} \\
\midrule
\multicolumn{7}{@{}l}{\textit{VLM}} \\
Claude Fable 5.2 & 60 & 28.58 & 27.37 & 35 & 12 & 13 \\
GPT-6.1-Sol & 60 & 22.28 & 20.96 & 38 & 13 & 9 \\
GPT-6-Astra & 60 & 20.06 & 18.70 & 39 & 13 & 8 \\
DeepSeek-V4.1-Flash & 60 & 11.86 & 10.37 & 48 & 7 & 5 \\
Kimi-K3.1-Pro & 60 & 9.08 & 7.54 & 49 & 8 & 3 \\
Qwen3.8-27B & 60 & 6.31 & 4.72 & 52 & 6 & 2 \\
Gemini-Decision-1 & 60 & 5.47 & 3.87 & 52 & 7 & 1 \\
Jev-1.14.0 & 60 & 4.64 & 3.02 & 53 & 6 & 1 \\
\multicolumn{7}{@{}l}{\textit{VLA}} \\
GR00T-N1.5 & 59 & 8.79 & 8.79 & 49 & 7 & 3 \\
$\pi_{0.5}$ (R1Pro SFT) & 59 & 8.11 & 8.11 & 50 & 6 & 3 \\
Xiaomi XR-1 & 59 & 2.82 & 2.82 & 54 & 5 & 0 \\
\multicolumn{7}{@{}l}{\textit{WAM}} \\
Awomo-0.5I & 59 & 6.75 & 6.75 & 51 & 6 & 2 \\
OpenWAM-Alpha & 59 & 3.81 & 3.81 & 52 & 7 & 0 \\
\bottomrule
\end{tabular}
\end{table}

\subsection{How Sol Changes Across Two Seeds}
\label{sec:sol_seed_repeat}

Sol's mean Interactive Score changes by less than a point across the two seeds, from 22.28 to 21.44. Its memory score rises from 27.08 to 41.67, while its physical reasoning score falls from 14.29 to zero. Sol completes the T21 tower only at seed~0 and the T46 change-guided delivery only at seed~1. Table~\ref{tab:sol_seed_repeat} gives the dimension scores on the same 60 tasks.

\begin{table}[H]
\centering\small
\setlength{\tabcolsep}{5pt}
\caption{Sol's dimension-level and overall Interactive Scores and overall terminal QA (0--100) on the same 60 tasks at seeds 0 and 1. The Mean row weights both seeds equally.}
\label{tab:sol_seed_repeat}
\begin{tabular}{@{}lrrrrrrrr@{}}
\toprule
Environment seed & Nav. & Ground. & Geom. & Phys. & Memory & Control & Overall Op. & Overall QA \\
\midrule
0 & 14.00 & 19.61 & 18.63 & 14.29 & 27.08 & 50.00 & 22.28 & 84.44 \\
1 & 17.33 & 25.49 & 13.73 & 0.00 & 41.67 & 33.33 & 21.44 & 86.94 \\
\midrule
Mean & 15.67 & 22.55 & 16.18 & 7.14 & 34.38 & 41.67 & 21.86 & 85.69 \\
\bottomrule
\end{tabular}
\end{table}

At seed 1, 40 tasks receive zero operation credit, 10 receive partial credit, and 10 receive full credit. At seed 0, the corresponding counts are 38, 13, and 9. Among episodes with full QA credit, zero operation credit occurs in 30 of 50 episodes at seed 0 and 32 of 51 at seed 1. Averaging over the 59 tasks shared by all evaluated models gives operation scores of 20.96 at seed 0 and 20.11 at seed 1.

\subsection{Model Interfaces and Episode Budgets}
\label{sec:analysis_settings}
\label{sec:episode_protocol}

Eight VLMs direct the R1 robot through WorldHarness. The five continuous-control policies use their trained robot interfaces. Table~\ref{tab:model_configurations} identifies each deployment and embodiment. Table~\ref{tab:policy_observations} gives the image histories and the number of control steps executed per action chunk.

\begin{table}[!htb]
\centering
\small
\setlength{\tabcolsep}{4pt}
\caption{Model deployments, reasoning settings, and robot interfaces for the cross-model evaluation. Vercel routes requests to the listed provider via AI Gateway. A dash indicates no configurable reasoning effort.}
\label{tab:model_configurations}
\begin{tabular}{@{}p{0.25\textwidth} p{0.23\textwidth} p{0.13\textwidth} p{0.19\textwidth} p{0.12\textwidth}@{}}
\toprule
\textbf{Model} & \textbf{Deployment} & \textbf{Reasoning}\newline\textbf{Effort} & \textbf{Harness} & \textbf{Robot} \\
\midrule
GPT-6.1-Sol & OpenAI (direct) & Medium & WorldHarness & R1 \\
GPT-6-Astra & Vercel / OpenAI & Medium & WorldHarness & R1 \\
Claude Fable 5.2 & Vercel / Anthropic & Medium & WorldHarness & R1 \\
Kimi-K3.1-Pro & Vercel / Moonshot AI & High & WorldHarness & R1 \\
DeepSeek-V4.1-Flash-1017 & Vercel / DeepSeek & High & WorldHarness & R1 \\
Qwen3.8-27B & Local, $1\times$ B200 & Medium & WorldHarness & R1 \\
Gemini-Decision-1 & Vercel / Google & Low & WorldHarness & R1 \\
Jev-1.14.0 & Vercel / Typesafe & --- & WorldHarness & R1 \\
\midrule
$\pi_{0.5}$ (R1Pro SFT) & Local, $1\times$ A5000  & --- & N/A & R1Pro \\
Xiaomi XR-1 & Local, $1\times$ A5000 & --- & N/A & PandaOmron \\
GR00T-N1.5 & Local, $1\times$ A5000 & --- & N/A & PandaOmron \\
Awomo-0.5I & Local, $1\times$ A5000 & --- & N/A & PandaOmron \\
OpenWAM-Alpha & Local, $1\times$ A5000 & --- & N/A & PandaOmron \\
\bottomrule
\end{tabular}
\end{table}

\noindent\textbf{Episode Budgets and World Timing.}
We evaluate all systems at environment seed~0 with task-specific budgets of 10--342 VLM decisions or 1,000--40,000 control steps. For the seven VLMs other than Qwen, world time continues to advance during inference on all 60 tasks. Qwen lets it advance on all 60. World time continues to advance during inference for all five continuous-control policies on all 59 evaluated tasks.

\noindent\textbf{VLM Context Settings.}
Sol, Astra, Fable, Kimi, DeepSeek, and Qwen choose parameterized commands using current observations, the last public result, and saved notes. Sol can retrieve earlier RGB views, make up to four rounds of perception calls, and save 128 characters per decision. Sol, Astra, Fable, and Kimi have input and output limits of 272,000 and 16,384 tokens. The output limits for Qwen and DeepSeek are 16,384 and 384,000 tokens. Qwen uses local vLLM with thinking enabled, temperature 1.0, and no guided decoding. Gemini and Jev use a fixed-choice interface of the kind studied in System One models~\cite{li2026jeva,li2026so}. From current views and public interaction records, they select a command and a repetition count of 1, 2, 4, or 8.

\noindent\textbf{Robot control.}
R1 uses base-position control, absolute-pose inverse kinematics for both arms, joint-position control for the trunk, and assisted grasp and release. Arm translations use the robot-base frame. PandaOmron policies run a 12-dimensional controller at 20\,Hz, with base velocity, torso control, arm operational-space control, and gripper joint positions. The R1Pro interface for $\pi_{0.5}$ combines base velocities and joint positions in 23 dimensions at 30\,Hz. All five continuous-control environments simulate physics at 120\,Hz.

\begin{table}[!htb]
\centering\small
\setlength{\tabcolsep}{7pt}
\caption{Observation and action settings for continuous-control policies. History counts observations, and Interval is their spacing in control steps. Source/Execute gives the source and executed action chunk lengths in control steps. Server reset records whether the adapter sent a BEHAVIOR reset message at initialization.}
\label{tab:policy_observations}
\begin{tabular}{@{}lrrrrl@{}}
\toprule
\textbf{Policy} & \textbf{Image size} & \textbf{History} & \textbf{Interval} & \textbf{Source/Execute} & \textbf{Server reset} \\
\midrule
$\pi_{0.5}$ & 224 & --- & --- & 26/20 & Not sent \\
XR-1 & 256 & 4 & 2 & 16/16 & Not sent \\
GR00T-N1.5 & 256 & 1 & 1 & 16/16 & Sent \\
Awomo-0.5I & 256 & 5 & 4 & 16/16 & Not sent \\
OpenWAM-Alpha & 256 & 1 & 1 & 32/32 & Not sent \\
\bottomrule
\end{tabular}
\end{table}

Awomo uses nine model-history slots spaced 40 control steps apart. The $\pi_{0.5}$ I/O interface accepts chunks up to 32 steps, while its action processor records 26 source steps and 20 executed steps.

\FloatBarrier
\section{Physical Outcomes Within an Episode}
\label{sec:episode_analysis}

\subsection{Physical Success and Goal Components}
\label{sec:episode_success}

Across 775 episodes, at least one system succeeds on 17 distinct tasks. One is T52, where all eight VLMs preserve an object that already lies in its container. On the 59 common tasks, Fable succeeds on 12, Sol on eight, Astra on seven, and DeepSeek on four. Table~\ref{tab:physical_success} breaks down these successes by the primary demand of each task. The evaluator checks 1,669 goal items across the 775 episodes.

\begin{table}[!htb]
\centering\small
\setlength{\tabcolsep}{6pt}
\caption{Physical success rates (\%) by dimension and overall at seed~0. Success requires all goals and constraints (Eq.~\ref{eq:physical_success}). The last column gives successes out of 60 tasks per VLM or 59 per continuous-control policy.}
\label{tab:physical_success}
\begin{tabular}{@{}lrrrrrrrr@{}}
\toprule
Model & Nav. & Ground. & Geom. & Phys. & Memory & Control & Overall & Successes \\
\midrule
Fable & 20.00 & 29.41 & 17.65 & 0.00 & 37.50 & 16.67 & 21.67 & 13/60 \\
Sol & 0.00 & 11.76 & 11.76 & 14.29 & 12.50 & 50.00 & 15.00 & 9/60 \\
Astra & 0.00 & 11.76 & 11.76 & 0.00 & 12.50 & 50.00 & 13.33 & 8/60 \\
DeepSeek & 0.00 & 5.88 & 11.76 & 0.00 & 12.50 & 16.67 & 8.33 & 5/60 \\
Kimi & 0.00 & 5.88 & 5.88 & 0.00 & 0.00 & 0.00 & 3.33 & 2/60 \\
Qwen & 0.00 & 0.00 & 5.88 & 0.00 & 12.50 & 0.00 & 3.33 & 2/60 \\
Gemini & 0.00 & 0.00 & 5.88 & 0.00 & 0.00 & 0.00 & 1.67 & 1/60 \\
Jev & 0.00 & 0.00 & 5.88 & 0.00 & 0.00 & 0.00 & 1.67 & 1/60 \\
GR00T-N1.5 & 0.00 & 11.76 & 0.00 & 0.00 & 12.50 & 0.00 & 5.08 & 3/59 \\
$\pi_{0.5}$ & 0.00 & 11.76 & 0.00 & 0.00 & 12.50 & 0.00 & 5.08 & 3/59 \\
Awomo-0.5I & 0.00 & 5.88 & 0.00 & 0.00 & 12.50 & 0.00 & 3.39 & 2/59 \\
OpenWAM & 0.00 & 0.00 & 0.00 & 0.00 & 0.00 & 0.00 & 0.00 & 0/59 \\
XR-1 & 0.00 & 0.00 & 0.00 & 0.00 & 0.00 & 0.00 & 0.00 & 0/59 \\
\bottomrule
\end{tabular}
\end{table}

\begin{figure}[!htbp]
\centering
\includegraphics[width=0.84\textwidth]{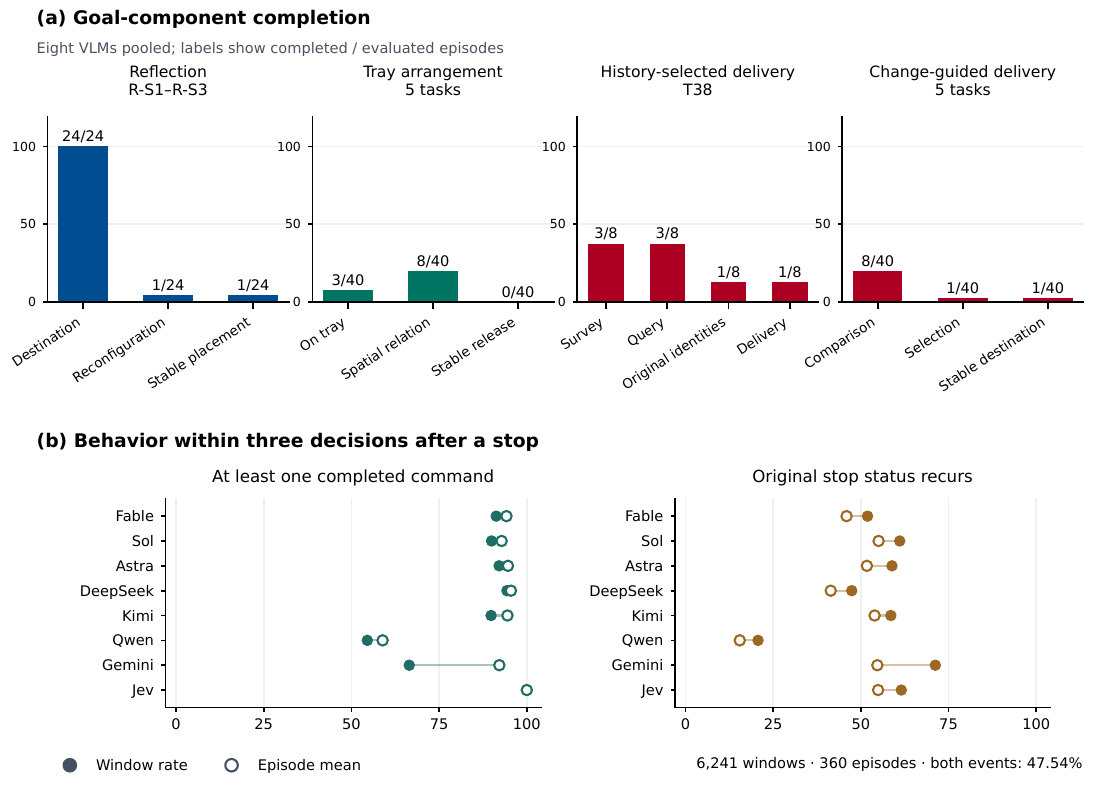}
\caption{(a) Goal-component completion by task family, pooled over eight VLMs. Bar labels give completed/evaluated episodes. (b) Shares of complete three-decision windows containing a completed command or a recurrence of the original stop status. Filled markers weight windows equally, and open markers average within each episode first.}
\label{fig:goals_feedback}
\end{figure}

\noindent\textbf{Destination, spatial relation, and stability.}
Of the 72 VLM episodes with partial credit, 23 come from the three mirror-guided rearrangements and 11 from the five tray arrangements (Fig.~\ref{fig:goals_feedback}a and Table~\ref{tab:goal_component_families}). Across all 24 mirror episodes the selected object reaches its destination. Only Fable on R-S1 also establishes the required spatial relation and stable placement. On T32 all eight VLMs satisfy the object-separation requirement, yet none places an object on the tray.

The destination check passes in two of three VLA runs on R-S1 and all three VLA runs on each of R-S2 and R-S3. All six WAM runs pass it. All 15 miss both the relation and the stability checks.

\noindent\textbf{Selecting an object from its history.}
Across T40 and T45--T48, eight of 40 VLM episodes complete the two-phase comparison. Fable on T46 then selects the changed object and leaves it stably at the specified destination. In T38, Fable, Sol, and Astra all complete the survey and reach the delayed request. Only Fable delivers the requested original objects (Table~\ref{tab:goal_components}). In the geometric assignment tasks, one of 16 episodes satisfies the assignment and none attains stable placement. The geometric passage and physical trial tasks receive zero component credit (Table~\ref{tab:goal_component_families}).

\noindent\textbf{Constraint Compliance.}
Kimi earns an Interactive Score of 100 for the T41 comparison. A task constraint fails. Only this episode among the 775 earns full component credit without physical success.

\begin{table}[!htb]
\centering\small
\setlength{\tabcolsep}{6pt}
\caption{Goal-component outcomes for Fable, Sol, and Astra on R-S1 and T38. The evaluator records 1 when a completion item is satisfied and 0 otherwise.}
\label{tab:goal_components}
\begin{tabular}{@{}llrrr@{}}
\toprule
Task & Completion item & Fable & Sol & Astra \\
\midrule
R-S1 & Selected object at destination & 1 & 1 & 1 \\
R-S1 & Requested spatial reconfiguration & 1 & 0 & 0 \\
R-S1 & Stable placement & 1 & 0 & 0 \\
T38 & Survey completed & 1 & 1 & 1 \\
T38 & Delayed query reached & 1 & 1 & 1 \\
T38 & Requested original identities & 1 & 0 & 0 \\
T38 & All requested items delivered & 1 & 0 & 0 \\
\bottomrule
\end{tabular}
\end{table}

\begin{table}[!htb]
\centering\small
\setlength{\tabcolsep}{6pt}
\caption{Goal-component completion in the final evaluator records, including checks on earlier events. Each row pools episodes within its task family, with one episode per VLM and task.}
\label{tab:goal_component_families}
\begin{tabular*}{\linewidth}{@{\extracolsep{\fill}}llrr@{}}
\toprule
Task family and IDs & Goal requirement & Completed / episodes & Rate (\%) \\
\midrule
\emph{Reflection reconfiguration} & Selected object at destination & 24/24 & 100.00 \\
{\footnotesize R-S1, R-S2, R-S3} & Requested reconfiguration & 1/24 & 4.17 \\
 & Stable placement & 1/24 & 4.17 \\
\addlinespace[3pt]
\emph{Tray arrangement} & Targets on tray & 3/40 & 7.50 \\
{\footnotesize T26, T28, T30, T32, T33} & Requested spatial relation & 8/40 & 20.00 \\
 & Stable released placement & 0/40 & 0.00 \\
\addlinespace[3pt]
\emph{History-selected delivery} & Survey completed & 3/8 & 37.50 \\
{\footnotesize T38} & Delayed query reached & 3/8 & 37.50 \\
 & Requested original identities & 1/8 & 12.50 \\
 & All requested items delivered & 1/8 & 12.50 \\
\addlinespace[3pt]
\emph{Change-guided delivery} & Two-phase comparison & 8/40 & 20.00 \\
{\footnotesize T40, T45, T46, T47, T48} & Correct object selected & 1/40 & 2.50 \\
 & Stable branch destination & 1/40 & 2.50 \\
\addlinespace[3pt]
\emph{Geometric passage} & Passage & 0/16 & 0.00 \\
{\footnotesize GF2-1, GF2-2} & Stable placement & 0/16 & 0.00 \\
\addlinespace[3pt]
\emph{Geometric assignment} & Assignment & 1/16 & 6.25 \\
{\footnotesize GF3-1, GF3-2} & Stable placement & 0/16 & 0.00 \\
\addlinespace[3pt]
\emph{Physical trials and delivery} & Controlled trials & 0/24 & 0.00 \\
{\footnotesize T42, T43, T44} & Evidence-based selection & 0/24 & 0.00 \\
 & Ramp delivery & 0/24 & 0.00 \\
\bottomrule
\end{tabular*}
\end{table}

\subsection{QA by Referenced State and History}
\label{sec:qa_episode_types}

\begin{figure}[htbp]
    \centering
    \includegraphics[width=0.78\linewidth]{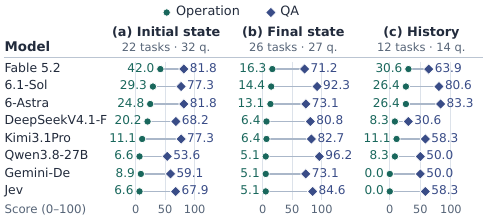}
    \caption{Mean paired operation and terminal QA by question type for the 22 tasks about the initial state, 26 about the final state, and 12 about history.}
    \label{fig:qa_by_question_type}
\end{figure}

The 73 terminal questions refer to the initial scene (32 questions across 22 tasks), the final state (27 across 26), or history (14 across 12). The history questions cover interaction events (eight questions) and changes between phases (six). We average question scores within tasks, then weight tasks equally within a group. Figure~\ref{fig:qa_by_question_type} and Table~\ref{tab:qa_types} pair these answers with physical progress. Table~\ref{tab:paired_qa_outcomes} counts physical outcomes when every answer is correct.

\begin{table}[!htb]
\centering\small
\setlength{\tabcolsep}{6pt}
\caption{Mean terminal QA and operation scores (0--100) from the same VLM rollouts. Tasks receive equal weight within the initial-state (22 tasks), final-state (26), and history (12) question groups.}
\label{tab:qa_types}
\begin{tabular}{@{}lrrrrrr@{}}
\toprule
Model & Initial QA & Initial Op. & Final QA & Final Op. & History QA & History Op. \\
\midrule
Fable & 81.82 & 41.97 & 71.15 & 16.35 & 63.89 & 30.56 \\
Sol & 77.27 & 29.32 & 92.31 & 14.42 & 80.56 & 26.39 \\
Astra & 81.82 & 24.77 & 73.08 & 13.14 & 83.33 & 26.39 \\
DeepSeek & 68.18 & 20.23 & 80.77 & 6.41 & 30.56 & 8.33 \\
Kimi & 77.27 & 11.14 & 82.69 & 6.41 & 58.33 & 11.11 \\
Qwen & 53.64 & 6.59 & 96.15 & 5.13 & 50.00 & 8.33 \\
Gemini & 59.09 & 8.86 & 73.08 & 5.13 & 50.00 & 0.00 \\
Jev & 67.88 & 6.59 & 84.62 & 5.13 & 58.33 & 0.00 \\
\bottomrule
\end{tabular}
\end{table}

\begin{table}[!htb]
\centering\small
\setlength{\tabcolsep}{7pt}
\caption{Mean QA uses all 60 tasks per VLM. Operation counts and zero-credit rates are calculated over the $N_{\mathrm{QA}}$ episodes with QA score 100.}
\label{tab:paired_qa_outcomes}
\begin{tabular}{@{}lrrrrrr@{}}
\toprule
\textbf{Model} & \textbf{Mean QA} & $N_{\mathrm{QA}}$ & \textbf{Zero} & \textbf{Partial} & \textbf{Full} & \textbf{Zero (\%)} \\
\midrule
Claude Fable 5.2 & 73.61 & 43 & 22 & 8 & 13 & 51.2 \\
GPT-6.1-Sol & 84.44 & 50 & 30 & 12 & 8 & 60.0 \\
GPT-6-Astra & 78.33 & 47 & 29 & 10 & 8 & 61.7 \\
DeepSeek-V4.1-Flash & 66.11 & 39 & 28 & 6 & 5 & 71.8 \\
Kimi-K3.1-Pro & 75.83 & 45 & 36 & 7 & 2 & 80.0 \\
Qwen3.8-27B & 71.33 & 42 & 36 & 5 & 1 & 85.7 \\
Gemini-Decision-1 & 63.33 & 38 & 30 & 7 & 1 & 78.9 \\
Jev-1.14.0 & 73.22 & 43 & 38 & 5 & 0 & 88.4 \\
\bottomrule
\end{tabular}
\end{table}

\noindent\textbf{What correct answers describe.}
On T42, Sol answers \texttt{none} when asked which candidate completed an eligible physical trial. The answer is correct and the physical score is zero. On T21, Sol builds an upright, stable tower and earns full physical credit, then answers the maximum-stability ordering question incorrectly.

\subsection{Decisions After a Motion Stops}
\label{sec:decision_statistics}

\begin{figure}[t]
\centering
\includegraphics[width=0.92\textwidth]{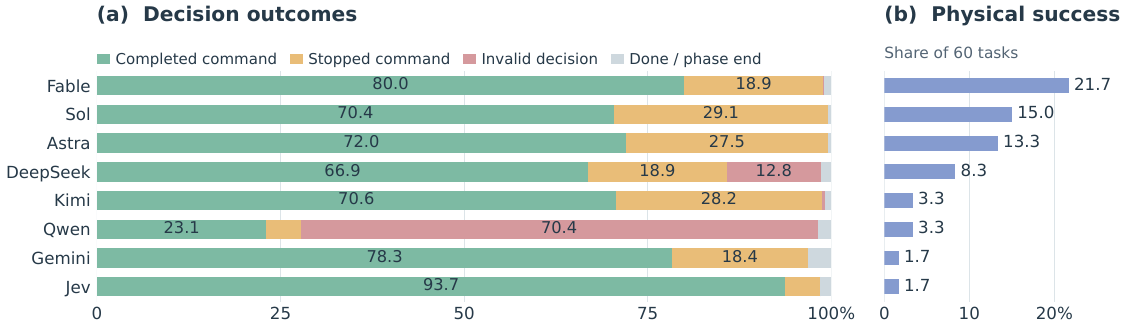}
\caption{Decision-level outcomes and task-level success for each VLM. The left panel uses all decisions, including invalid decisions and phase or final declarations. The right panel reports the percentage of 60 tasks that meet all physical goals and constraints.}
\label{fig:decision_outcomes}
\end{figure}

WorldHarness records 32,455 decisions. Table~\ref{tab:decision_counts} separates invalid decisions, phase or final declarations, and motions the controller executes. An executed motion either completes or stops, for example at a joint limit or a blocked path. Figure~\ref{fig:decision_outcomes} places command outcomes beside each VLM's task success rate.

\begin{table}[!htb]
\centering\small
\setlength{\tabcolsep}{6pt}
\caption{Counts of all recorded decisions by VLM. Declarations include phase endings and final \texttt{DONE}. Completed (\%) uses executed commands as its denominator (Completed + Stopped).}
\label{tab:decision_counts}
\begin{tabular}{@{}lrrrrrr@{}}
\toprule
Model & Decisions & Completed & Stopped & Invalid & Declarations & Completed (\%) \\
\midrule
Fable & 4,152 & 3,320 & 786 & 5 & 41 & 80.86 \\
Sol & 4,779 & 3,365 & 1,392 & 0 & 22 & 70.74 \\
Astra & 4,676 & 3,368 & 1,287 & 0 & 21 & 72.35 \\
DeepSeek & 4,396 & 2,940 & 831 & 561 & 64 & 77.96 \\
Kimi & 5,011 & 3,539 & 1,411 & 18 & 43 & 71.49 \\
Qwen & 4,440 & 1,025 & 206 & 3,127 & 82 & 83.27 \\
Gemini & 2,129 & 1,668 & 392 & 0 & 69 & 80.97 \\
Jev & 2,872 & 2,692 & 134 & 0 & 46 & 95.26 \\
\bottomrule
\end{tabular}
\end{table}

\noindent\textbf{The Next Decision after a Stopped Command.}
Sol repeats both action and parameters immediately after a stop in 13 of 1,382 pairs. Astra does so once in 1,278. Most next decisions select a different action, and others change the parameters. Qwen produces an invalid decision or a declaration in 158 of its 206 pairs (Table~\ref{tab:feedback_choices}).

\begin{table}[!htb]
\centering\small
\setlength{\tabcolsep}{6pt}
\caption{Counts of next decisions after a stopped command, for stops with a following decision in the episode. Same command repeats the action name and parameters. Changed parameters keeps the action name, and Changed action uses a different name.}
\label{tab:feedback_choices}
\begin{tabular}{@{}lrrrrr@{}}
\toprule
Model & Pairs & Same command & Changed parameters & Changed action & Invalid / declaration \\
\midrule
Fable & 782 & 27 & 38 & 711 & 6 \\
Sol & 1382 & 13 & 37 & 1329 & 3 \\
Astra & 1278 & 1 & 15 & 1258 & 4 \\
DeepSeek & 830 & 1 & 15 & 771 & 43 \\
Kimi & 1400 & 4 & 24 & 1369 & 3 \\
Qwen & 206 & 0 & 2 & 46 & 158 \\
Gemini & 391 & 39 & 39 & 313 & 0 \\
Jev & 133 & 0 & 2 & 131 & 0 \\
\bottomrule
\end{tabular}
\end{table}

\noindent\textbf{The next three decisions.}
Of 6,439 stopped commands, 6,241 have three subsequent decisions in the same episode. These windows span 360 episodes and include executed motions, invalid decisions, and declarations. We count whether any subsequent motion completes and whether the original stop status recurs. Both occur in 2,967 windows (47.54\%). Table~\ref{tab:feedback_followup} also gives the completion rate on the next decision.

Arm blockage is the most common stop. In 64.57\% of the 3,458 windows after arm blockage, arm blockage recurs. Across stop types, 24.72\% of windows repeat the original action name and stop status.

\begin{table}[!htb]
\centering\small
\setlength{\tabcolsep}{5pt}
\caption{Rates (\%) over complete three-decision windows after a stopped command. Next measures completion on the immediate next decision, and Within 3 measures at least one completion in the window. Repeat measures recurrence of the original stop status. Episode means weight eligible episodes equally.}
\label{tab:feedback_followup}
\begin{tabular}{@{}lrrrrrrr@{}}
\toprule
 & & & \multicolumn{3}{c}{Window rates} & \multicolumn{2}{c}{Episode means} \\
\cmidrule(lr){4-6}\cmidrule(lr){7-8}
Model & Windows & Episodes & Next & Within 3 & Repeat & Within 3 & Repeat \\
\midrule
Fable & 757 & 56 & 62.22 & 91.28 & 51.92 & 94.21 & 45.94 \\
Sol & 1,344 & 58 & 56.40 & 89.96 & 61.09 & 92.83 & 55.07 \\
Astra & 1,255 & 55 & 59.04 & 92.11 & 58.88 & 94.63 & 51.74 \\
DeepSeek & 812 & 54 & 64.04 & 94.33 & 47.41 & 95.48 & 41.42 \\
Kimi & 1,366 & 58 & 58.86 & 89.82 & 58.57 & 94.50 & 53.93 \\
Qwen & 198 & 37 & 15.66 & 54.55 & 20.71 & 58.91 & 15.50 \\
Gemini & 379 & 33 & 43.54 & 66.49 & 71.24 & 92.18 & 54.71 \\
Jev & 130 & 9 & 87.69 & 100.00 & 61.54 & 100.00 & 54.93 \\
\midrule
All VLMs & 6,241 & 360 & 57.75 & 88.75 & 56.55 & 90.62 & 46.81 \\
\bottomrule
\end{tabular}
\end{table}

\noindent\textbf{Window length and episode weighting.}
Giving each eligible episode equal weight, 90.62\% of three-decision windows contain a completed command and 46.81\% repeat the original stop status. With five subsequent decisions, the corresponding episode means are 95.87\% and 58.40\%. Pooling the 6,034 five-decision windows gives 94.95\% and 67.88\%, respectively.

\subsection{Three Routes to Physical Success}
\label{sec:trajectory_evidence}

Figure~\ref{fig:physical_trajectories} follows three successful episodes, one per track. Table~\ref{tab:trajectory_commands} links their commands to the frame indices. The physical checks below identify the outcome each robot leaves behind.

\begin{table}[!htb]
\centering\small
\setlength{\tabcolsep}{6pt}
\caption{Selected decisions from the three episodes in Fig.~\ref{fig:physical_trajectories}, with commands and execution results.}
\label{tab:trajectory_commands}
\begin{tabular}{@{}lrll@{}}
\toprule
Episode & Decision & Command & Execution result \\
\midrule
Fable / R-S1 & 38 & \texttt{GRASP} & Completed \\
Fable / R-S1 & 40 & \texttt{TRUNK\_UP}, 0.10\,m & Completed \\
Fable / R-S1 & 45 & \texttt{RELEASE} & Completed \\
Sol / T21 & 40 & \texttt{MV\_UP}, 0.08\,m & Arm joint limit \\
Sol / T21 & 63 & \texttt{MV\_UP}, 0.08\,m & Completed \\
Sol / T21 & 75 & \texttt{RELEASE} & Completed \\
Astra / \texttt{iw\_h2\_02} & 36 & \texttt{BASE\_RIGHT}, 0.40\,m & Completed \\
Astra / \texttt{iw\_h2\_02} & 49 & \texttt{RELEASE} & Completed \\
\bottomrule
\end{tabular}
\end{table}

\noindent\textbf{Reflection-Guided Reconfiguration.}
Fable releases the selected object at decision 45 and ends the episode at decision 47. The object is at its destination, in the requested relation to the pot, and stable for 11.53 seconds. The top row of Fig.~\ref{fig:physical_trajectories} follows its approach through withdrawal.

\noindent\textbf{Stable Stacking.}
Sol leaves the chopping board at the bottom, the cookie box in the middle, and the shoe box on top. After releasing the first box it encounters an arm joint limit at decision 40. At termination (decision 79), contact, uprightness, stationary motion, and free grippers all hold over 31 samples spanning one second.
In the successful run, Sol first secures the cookie box at decision 27 after two attempts, raising it about 8\,cm. It releases the box at decision 39 and the shoe box at decision 75. In the pre-action QA run on the same task, Sol first attaches the cookie box at decision 94 after 12 grasp attempts. It releases the box at decision 106 and reaches the decision limit with the shoe box still at its starting position.

\noindent\textbf{Disturbance Repair.}
Astra clears the obstruction, delivers the apple, and observes the returning obstruction at decision 36. By its final declaration at decision 55 it has stored the obstruction and left the work area clear with protected objects preserved. The episode spans 115.43 simulated seconds and includes three releases.

\subsection{When Agents End Interaction}
\label{sec:episode_termination}

Episodes terminate at the final completion declaration or the interaction budget. Intermediate phase declarations contribute to the declaration count in Table~\ref{tab:decision_counts}; Table~\ref{tab:episode_stops} counts only final outcomes. Among 279 final declarations, 41 episodes achieve physical success, 196 receive zero credit, 41 receive partial credit, and one receives full component credit with an unmet constraint. The 201 budget-terminated episodes comprise 170 zero-credit and 31 partial-credit outcomes. All 41 successful VLM episodes end with a final declaration, including the eight T52 containment controls.

\begin{figure}[htbp]
\centering
\includegraphics[width=0.90\textwidth]{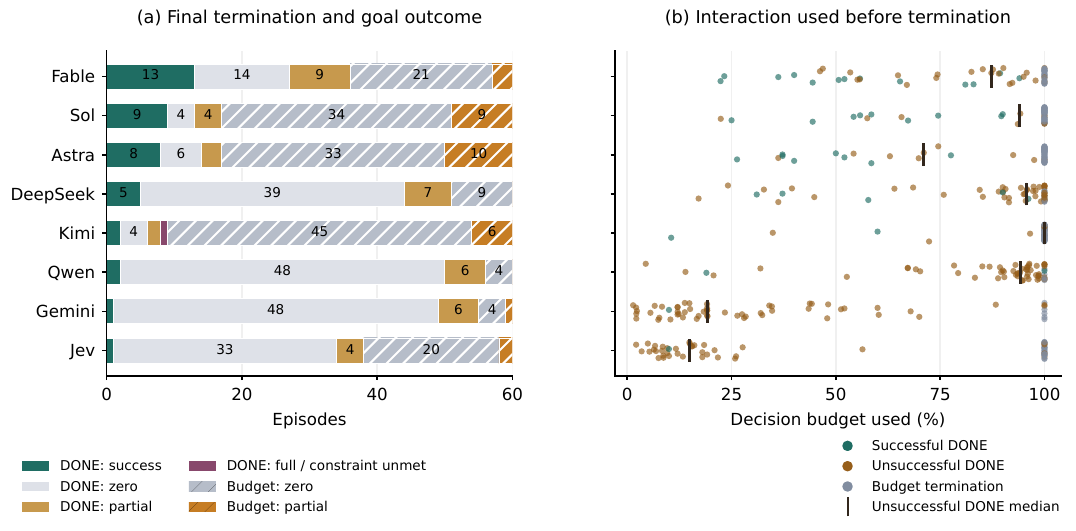}
\caption{Termination across 60 episodes per VLM. (a) Physical outcomes after final \texttt{DONE} or budget termination (hatched). (b) Each dot shows the share of the decision budget used in one episode. Dark ticks mark the median budget share for unsuccessful final declarations.}
\label{fig:termination}
\end{figure}

\noindent\textbf{Budget Use at the Final Declaration.}
Gemini and Jev use medians of 19.15\% and 14.89\% of their budgets before an unsuccessful declaration. Qwen, DeepSeek, and Sol use 94.36\%, 95.79\%, and 93.92\%, respectively (Fig.~\ref{fig:termination}). A last-decision declaration counts as a final \texttt{DONE}, as in 41 of the 279 declared episodes. Sol and Astra each have 43 more episodes that exhaust the budget, with mean Interactive Scores of 6.67 and 7.05.

\begin{table}[!htb]
\centering\footnotesize
\setlength{\tabcolsep}{4pt}
\caption{VLM episodes by termination and physical outcome. Zero and Partial denote $I=0$ and $0<I<100$. Full$^{\ast}$ denotes $I=100$ with an unmet constraint. Success requires all goals and constraints. Used is the median budget share for unsuccessful final declarations. Op.\ is the mean Interactive Score for episodes that end at the interaction budget.}
\label{tab:episode_stops}
\begin{tabular}{@{}lrrrrrrrrrr@{}}
\toprule
 & \multicolumn{6}{c}{Final declaration} & \multicolumn{4}{c}{Budget termination} \\
\cmidrule(lr){2-7}\cmidrule(lr){8-11}
Model & Total & Zero & Partial & Full$^{\ast}$ & Success & Used (\%) & Total & Zero & Partial & Op. \\
\midrule
Fable & 36 & 14 & 9 & 0 & 13 & 87.27 & 24 & 21 & 3 & 3.47 \\
Sol & 17 & 4 & 4 & 0 & 9 & 93.92 & 43 & 34 & 9 & 6.67 \\
Astra & 17 & 6 & 3 & 0 & 8 & 71.07 & 43 & 33 & 10 & 7.05 \\
DeepSeek & 51 & 39 & 7 & 0 & 5 & 95.79 & 9 & 9 & 0 & 0.00 \\
Kimi & 9 & 4 & 2 & 1 & 2 & 100.00 & 51 & 45 & 6 & 3.92 \\
Qwen & 56 & 48 & 6 & 0 & 2 & 94.36 & 4 & 4 & 0 & 0.00 \\
Gemini & 55 & 48 & 6 & 0 & 1 & 19.15 & 5 & 4 & 1 & 5.00 \\
Jev & 38 & 33 & 4 & 0 & 1 & 14.89 & 22 & 20 & 2 & 2.05 \\
\midrule
All VLMs & 279 & 196 & 41 & 1 & 41 & 74.23 & 201 & 170 & 31 & 4.69 \\
\bottomrule
\end{tabular}
\end{table}

\subsection{Sol Under Changes to Its Interface}
\label{sec:sol_diagnostics}

All ablations use GPT-6.1-Sol on R1 with medium reasoning effort and scene seed 0, with one episode per task and condition. The QA timing and evidence ablation uses the direct OpenAI endpoint. The action-abstraction, feedback, and notes ablations use Vercel AI Gateway, and world time continues to advance during inference in all ablations. Table~\ref{tab:sol_diagnostic_scores} reports operation and QA scores for these three ablations on the main 0--100 scales.

\noindent\textbf{QA Timing and Evidence.}
For each of 12 selected tasks, we use one question with a fixed reference answer. The pre-action request sees the initial head and wrist images, and its answer is kept out of the controller context. Terminal QA uses the initial overview, final observations, and interaction notes. Three of six tasks with a correct pre-action answer end in physical success; the other three receive zero operation credit. Two of six tasks with an incorrect pre-action answer end in physical success. Four answers change from incorrect to correct at termination.

\begin{table}[!htb]
\centering\small
\setlength{\tabcolsep}{8pt}
\caption{Sol's operation and terminal QA scores (0--100) on the selected ablation tasks, with one episode per task and condition at seed~0. Each intervention is paired with a reference using motion commands, full feedback, and visible notes.}
\label{tab:sol_diagnostic_scores}
\begin{tabular}{@{}llrrrr@{}}
\toprule
& & \multicolumn{2}{c}{Reference} & \multicolumn{2}{c}{Intervention} \\
\cmidrule(lr){3-4}\cmidrule(l){5-6}
Intervention & Task & Operation & QA & Operation & QA \\
\midrule
Add object skills & T18 & 0 & 0 & 0 & 0 \\
& T21 & 0 & 0 & 0 & 0 \\
& T24 & 100 & 100 & 0 & 100 \\
\addlinespace
Hide execution feedback & R-S1 & 33.33 & 100 & 33.33 & 100 \\
& T22 & 100 & 100 & 0 & 100 \\
\addlinespace
Hide saved notes & \texttt{iw\_m2\_01} & 0 & 50 & 0 & 50 \\
\bottomrule
\end{tabular}
\end{table}

\noindent\textbf{Action Abstraction.}
We add named grasp and placement skills to the motion interface, with shared scenes, resets, cameras, and a nominal 5,000-control budget. Four runs record 5,003 controller steps. Sol chooses targets and positions the robot. The skill executor uses measured object bounds for local arm and gripper motions. Of 52 named grasps, 30 targets fall outside the workspace, 19 approaches stop before grasping, two commands are cancelled, and one fails the orientation check. The T24 motion baseline succeeds in 3,700 controls; the skill condition receives zero operation credit. Both interfaces receive zero operation credit on T18 and T21.

\noindent\textbf{Execution Feedback.}
The intervention removes execution status, reasons, and recovery hints from the model's decision context. Images, robot state, commands, and motion measurements remain available. On T22, full feedback gives verified placement and a completion declaration at decision 41. The masked run reaches its 52-decision limit with placement unverified. The shared terminal QA protocol gives full credit in both, so the physical score captures the difference in achieved state.

\noindent\textbf{Saved Notes.}
Both runs retain action receipts and current and historical RGB access. The intervention withholds saved notes during decisions and terminal QA. Each run satisfies three of four placement checks, missing different items. Sol declares completion at decision 74 with notes and uses all 94 with notes hidden.

\subsection{Physical Conditions for Success}
\label{sec:episode_validation}

Reference runs meet the fixed suite's goals and constraints. T52 checks initial containment. Model episodes undergo the same checks on object geometry, motion, contacts, and time-stamped events.

Task constraints specify the objects to preserve, event order, and terminal conditions.

\noindent\textbf{Insertion Conditions.}
For T22--T25, partial insertion is measured against the container's current pose, using its interior and solid geometry. Container contact is optional. The released target must remain stable for 0.5\,s, with both hands empty and at least 2\,cm clear, and the working gripper open. Containers and supports may move or rotate while object scales remain fixed. Targets may be verified sequentially and removed after verification to free a shared container.

\begin{table}[!htb]
\centering\small
\setlength{\tabcolsep}{9pt}
\caption{Evaluator thresholds for insertion and stable, released placement in tasks T22--T25.}
\label{tab:insertion_thresholds}
\begin{tabular}{@{}lr@{}}
\toprule
Quantity & Threshold \\
\midrule
Stable interval & 0.5\,s \\
Maximum linear speed & 0.02\,m/s \\
Maximum angular speed & 0.1\,rad/s \\
Maximum gap between samples & 0.1\,s \\
Interior margin & 2\,mm \\
Solid-geometry penetration tolerance & 1\,mm \\
Hand withdrawal clearance & 2\,cm \\
\bottomrule
\end{tabular}
\end{table}

\noindent\textbf{Other Task Thresholds.}
Reflection-guided placement uses a 0.5\,s stable interval, maximum linear speed of 0.02\,m/s, maximum angular speed of 0.1\,rad/s, and a maximum sample gap of 0.1\,s. R-S1 additionally requires the selected object to be at least 20\,cm farther from the fixed pot than the other object. T42--T44 use 3\,s trial intervals and treat arrival times within 0.15\,s as tied.

\FloatBarrier
\section{The Cost of Physical Interaction}
\label{sec:resource_appendix}

Table~\ref{tab:resource_full} separates model requests, simulator execution, and total elapsed time, which includes observations and harness overhead. Figure~\ref{fig:score_efficiency} in the main text compares physical progress with elapsed time and token use.

Navigation takes the most inference time per task for each of the 13 systems (Table~\ref{tab:dimension_inference_time}).

\noindent\textbf{Which calls take the time?}
Terminal QA accounts for 0.10--2.70\% of VLM inference time. Sol averages 124.80 interaction requests and one terminal request per task. For DeepSeek those means are 257.78 and one. Table~\ref{tab:phase_usage} separates interaction from terminal QA.

\begin{table}[H]
\centering
\small
\setlength{\tabcolsep}{4pt}
\renewcommand{\arraystretch}{1.12}
\caption{Resource use by model. Input, output, and total tokens are suite totals in millions. Mean tokens, API cost (USD), and time (min) are per task. VLM totals include terminal QA. The total time also accounts for observation and harness overhead.}
\label{tab:resource_full}
\label{tab:token_usage}
\begin{tabular*}{\textwidth}{@{}l@{\extracolsep{\fill}}rrrrrrrr@{}}
\toprule
 & \multicolumn{5}{c}{\cellcolor[HTML]{EBEDF3}\textbf{Tokens and API cost}} & \multicolumn{3}{c}{\cellcolor[HTML]{EBEDF3}\textbf{Time (min/task)}} \\
\cmidrule(lr){2-6}\cmidrule(lr){7-9}
\textbf{Model} & Input & Output & Total & Mean/task & Cost/task & Inference & Simulator & Total \\
\midrule
\multicolumn{9}{@{}>{\columncolor[HTML]{DEDEDE}[0pt][0pt]}l@{}}{\textbf{\textit{VLM}}} \\
Jev & 70.68 & 4.67 & 75.35 & 1.26 & 0.05 & 1.92 & 5.26 & 12.53 \\
Gemini-Decision & 46.46 & 0.00 & 46.46 & 0.77 & 0.08 & 3.27 & 4.63 & 11.66 \\
6.1-Sol & 145.14 & 0.98 & 146.13 & 2.44 & 5.00 & 14.74 & 18.92 & 46.36 \\
6-Astra & 134.12 & 0.95 & 135.07 & 2.25 & 23.15 & 16.33 & 18.48 & 40.80 \\
Fable 5.2 & 179.54 & 4.47 & 184.00 & 3.07 & 26.92 & 26.81 & 16.86 & 51.12 \\
Qwen3.8-27B & 375.62 & 12.74 & 388.37 & 6.47 & --- & 39.22 & 4.45 & 51.05 \\
Kimi3.1Pro & 186.69 & 3.90 & 190.59 & 3.18 & 6.87 & 51.47 & 21.87 & 90.70 \\
DeepseekV4.1-F & 354.41 & 78.69 & 433.09 & 7.22 & 1.67 & 133.12 & 16.41 & 171.58 \\
\multicolumn{9}{@{}>{\columncolor[HTML]{DEDEDE}[0pt][0pt]}l@{}}{\textbf{\textit{VLA}}} \\
GR00T-N1.5 & --- & --- & --- & --- & --- & 1.08 & 16.39 & 23.83 \\
$\pi_{0.5}$ & --- & --- & --- & --- & --- & 1.38 & 13.06 & 30.41 \\
XiaomiXR1 & --- & --- & --- & --- & --- & 5.30 & 19.50 & 31.69 \\
\multicolumn{9}{@{}>{\columncolor[HTML]{DEDEDE}[0pt][0pt]}l@{}}{\textbf{\textit{WAM}}} \\
OpenWAM$\alpha$ & --- & --- & --- & --- & --- & 5.52 & 16.02 & 27.66 \\
Awomo-0.5I & --- & --- & --- & --- & --- & 19.62 & 35.28 & 62.51 \\
\bottomrule
\end{tabular*}
\end{table}

\begin{table}[H]
\centering\small
\setlength{\tabcolsep}{4pt}
\caption{Mean inference minutes per task by primary dimension at seed~0. The time spent on terminal QA is included in each VLM's request time.}
\label{tab:dimension_inference_time}
\begin{tabular}{@{}lrrrrrrr@{}}
\toprule
System & Navigation & Grounding & Geometry & Physics & Memory & Constraints & Overall \\
\midrule
Jev-1.14.0 & 4.55 & 1.67 & 1.51 & 1.82 & 1.38 & 2.44 & 1.92 \\
Gemini-Decision-1 & 15.75 & 2.12 & 1.61 & 4.09 & 2.27 & 1.15 & 3.27 \\
Qwen3.8-27B & 87.17 & 41.29 & 37.13 & 29.39 & 21.81 & 34.00 & 39.22 \\
Kimi-K3.1-Pro & 133.71 & 37.65 & 44.90 & 44.37 & 45.08 & 57.51 & 51.47 \\
DeepSeek-V4.1-Flash & 264.77 & 129.70 & 128.93 & 99.05 & 77.57 & 158.75 & 133.12 \\
GPT-6-Astra & 52.09 & 11.17 & 13.17 & 15.82 & 15.36 & 11.98 & 16.33 \\
GPT-6.1-Sol & 38.87 & 10.64 & 12.26 & 15.56 & 14.64 & 12.42 & 14.74 \\
Claude Fable 5.2 & 52.79 & 24.44 & 23.98 & 27.37 & 24.67 & 22.07 & 26.81 \\
\midrule
Xiaomi XR-1 & 13.49 & 4.63 & 4.93 & 5.58 & 3.63 & 3.24 & 5.30 \\
$\pi_{0.5}$ (R1Pro SFT) & 3.53 & 1.19 & 1.33 & 1.52 & 0.83 & 0.83 & 1.38 \\
GR00T-N1.5 & 2.73 & 0.92 & 1.02 & 1.15 & 0.77 & 0.64 & 1.08 \\
\midrule
OpenWAM-Alpha & 9.33 & 4.46 & 4.87 & 4.87 & 8.47 & 3.92 & 5.52 \\
Awomo-0.5I & 51.81 & 16.12 & 17.91 & 21.24 & 15.35 & 11.09 & 19.62 \\
\bottomrule
\end{tabular}
\end{table}

\begin{table}[H]
\centering\small
\setlength{\tabcolsep}{6pt}
\caption{Mean API calls and inference minutes per task, split into interaction and terminal QA. Call counts include request attempts and tool rounds. QA share is the percentage of the total recorded inference time that each model spends answering the terminal questions.}
\label{tab:phase_usage}
\begin{tabular}{@{}lrrrrr@{}}
\toprule
Model & Interaction calls & QA calls & Interaction time & QA time & QA share (\%) \\
\midrule
Fable & 92.80 & 1.00 & 26.62 & 0.19 & 0.70 \\
Sol & 124.80 & 1.00 & 14.64 & 0.09 & 0.64 \\
Astra & 119.00 & 1.00 & 16.21 & 0.12 & 0.71 \\
DeepSeek & 257.78 & 1.00 & 132.97 & 0.13 & 0.10 \\
Kimi & 139.32 & 1.00 & 51.26 & 0.21 & 0.41 \\
Qwen & 285.75 & 2.00 & 39.12 & 0.10 & 0.26 \\
Gemini & 35.62 & 1.03 & 3.18 & 0.09 & 2.70 \\
Jev & 47.88 & 1.00 & 1.88 & 0.04 & 2.16 \\
\bottomrule
\end{tabular}
\end{table}

\linespread{1.04}\selectfont
\setlength{\parskip}{0.73\baselineskip}
\section{Scene Generation}
\label{sec:dimension_generation}

A new layout changes where the robot must go and what it can reach. To preserve the task, the generator holds its goal, constraints, and physical predicates fixed and varies the scene, objects, starting pose, or events. For example, a narrower slot may require a new approach, but the reoriented board must still fit.

\subsection{Which Instances Preserve the Goal?}

For a task template $i$ in dimension $d$, let $z=(e,o,x,r,\eta)$ contain the scene, selected objects, object poses, robot starting pose, and task-event parameters. The generator draws proposals from $q_i$. Initialization conditions $V_i(z)$, dimension-specific challenge conditions $K_d(z)$, and manual review $H_i(z)$ define which proposals are accepted. The resulting conditional distribution is
\begin{equation}
 p_i(z)=\frac{q_i(z)\,V_i(z)\,K_d(z)\,H_i(z)}{Z_i},
 \label{eq:instance_generation}
\end{equation}
where the three acceptance functions take values in $\{0,1\}$ and $Z_i>0$ is the normalizing constant. The interface exposes scene choice, object sizes, layout extent, robot-start regions, event parameters, and seed. Sampling takes a target instance count and a finite candidate budget.

The initialization conditions cover role-compatible objects, object dimensions, overlap, support placement, and navigation and manipulation access. The dimension conditions preserve each template's principal challenge. Manual review checks instruction clarity, feasible interactions, the assigned capability, and unintended shortcuts. The review outcome is $H_i(z)$.

\subsection{WorldSeek Routes and Evidence}

\noindent\textbf{Spatial navigation.}
Represent traversable regions by a graph $G_z=(V_z,E_z)$, with edges for robot-accessible passages. Sample a starting region, goal regions, visit order, and object placements within those regions. Layout variation changes obstacle positions and route lengths while retaining a path satisfying the required visit sequence. For a route $\gamma$, let $L(\gamma)$ denote the ordered regions visited and let $\sigma_i$ be the required sequence. A candidate satisfies the route condition when
\begin{equation}
 \exists\gamma\subseteq G_z:\quad \sigma_i\preceq L(\gamma),
 \label{eq:generate_navigation}
\end{equation}
where $\preceq$ denotes an ordered subsequence. A connectivity template preserves the adjacency relation queried by its goal. A collection template additionally preserves access to every required object and destination. The template varies route length, branching, and goal-region separation. Manual screening checks that the passage geometry admits the required route and that the task requires the intended region or route reasoning.

\noindent\textbf{Goal grounding.}
Sample a target, distractors, task cues, and their spatial arrangement. Occluders, inspection surfaces, or reflectors are placed according to the template. Let $C(h)$ be the set of target candidates consistent with the task cues observed in history $h$. For templates requiring disambiguation, construct an initial history $h_0$ and an accessible observation sequence $a_{1:k}$ such that
\begin{equation}
 |C(h_0)|>1,\qquad C(h_0,a_{1:k},o_{1:k})=\{o^\star\},
 \label{eq:generate_grounding}
\end{equation}
where $o^\star$ is the intended target. For direct grounding templates, the condition instead requires an observable cue that uniquely identifies the target. Sample distractor similarity, occlusion, and cue-viewpoint separation within these conditions. Manual screening checks cue visibility, the correspondence between the cue and the real object, and access to the required observation and retrieval actions.

\subsection{WorldShape Geometry and Physical Outcomes}

\noindent\textbf{Geometric reconfiguration.}
Sample object dimensions and poses jointly with apertures, receiving regions, or support geometry. Let $O$ denote an object's occupied volume, $R$ its orientation, and $b(O,R)$ its projected extents. For an orientation-sensitive opening with usable dimensions $a$, choose an initial orientation $R_0$ and a feasible orientation $R^\star$ satisfying
\begin{equation}
 b(O,R^\star)+m\mathbf{1}\leq a,
 \qquad b(O,R_0)+m\mathbf{1}\nleq a,
 \label{eq:generate_geometry}
\end{equation}
with clearance margin $m>0$. The instance conditions also require a collision-free approach and insertion path. Arrangement and sorting templates sample objects whose dimensions or relative poses determine distinct goal configurations. Sampling varies clearances, orientation changes, and competing placements. Review checks access, graspability, and the required reconfiguration. The T52 control template preserves its initially satisfied containment goal.

\noindent\textbf{Physical reasoning.}
Sample object shapes, support arrangements, and the physical parameters permitted by the template. Represent a candidate action sequence by $u$ and its predicted outcome under the template's physical model by $Y_z(u)$. Construct a feasible sequence $u^+$ and a competing sequence $u^-$ with
\begin{equation}
 G_i(Y_z(u^+))=1,\qquad G_i(Y_z(u^-))=0,
 \label{eq:generate_physics}
\end{equation}
where $G_i$ denotes satisfaction of the template's physical goal. In stacking tasks, contact geometry and the projected center of mass relative to the support region describe the stability challenge. Ramp-trial templates vary candidate shapes and release conditions while holding the comparison conditions fixed. Review uses simulator trajectories to examine contact, release, settling, and the physical outcomes of different actions.

\subsection{WorldShift Memory and Recovery}

\noindent\textbf{Temporal memory.}
Sample object identities, an observation phase, intervening events, and a later instruction that depends on the observed history. Events can move, exchange, remove, or replace objects within the template's rules. For a history-dependent identity task, construct two possible histories $h$ and $h'$ with the same current task-relevant observation and different required targets, so that
\begin{equation}
 o_t(h)=o_t(h'),\qquad y_i(h)\neq y_i(h'),
 \label{eq:generate_memory}
\end{equation}
where $y_i$ denotes the correct target or response. Event-tracking templates preserve the required order of observations and changes. Sampling varies the number of relevant objects, intervening changes, and delay. Manual screening checks that earlier observations reveal the needed information, object identity remains coherent through the event sequence, and the later goal requires the intended historical information.

\noindent\textbf{Constraint management.}
Sample resource quantities, compatible replacements, protected objects, delivery regions, and disturbances. For state $s$, define the set of feasible plans
\begin{equation}
 \Pi_i(s)=\{\pi:G_i(\tau_{s,\pi})=1,\ C_i(\tau_{s,\pi})=1\},
 \label{eq:generate_recovery}
\end{equation}
where $G_i$ and $C_i$ aggregate the goal and constraint predicates and $\tau_{s,\pi}$ is the resulting trajectory. A recovery template samples a disturbance $\delta$ such that $\Pi_i(\delta(s))$ is nonempty and the previous plan requires a new continuation. Resource-allocation templates preserve at least one feasible assignment while making resource or compatibility constraints consequential. Manual screening checks disturbance observability, access to recovery actions, and compliance with the declared constraints.

\end{document}